\documentclass[twoside]{article}

\usepackage{microtype}
\usepackage[accepted]{aistats2020}

\makeatletter
\renewcommand{\Notice@String}{\textsuperscript{$\dagger$} %
Contributed during an internship at QuantumBlack. \AISTATS@appearing}
\makeatother

\usepackage[round]{natbib}

\usepackage{amsmath,amssymb,bm,mathtools}
\usepackage{xr}
\usepackage[hidelinks]{hyperref}
\usepackage{cleveref}
\usepackage{graphicx,subcaption}
\usepackage{xcolor}
\usepackage{todonotes}
\usepackage{booktabs}
\usepackage{enumitem}
\usepackage[normalem]{ulem}
\usepackage{appendix}
\usepackage{amsthm}
\usepackage{placeins}
\usepackage{microtype}

\newcommand{\mat}[1]{\bm{\mathrm{#1}}}
\renewcommand{\vec}[1]{\bm{\mathrm{#1}}}

\newcommand{\Amat}{\mat{A}}
\newcommand{\Bmat}{\mat{B}}
\newcommand{\emat}{\mat{e}}
\newcommand{\Imat}{\mat{I}}
\newcommand{\Wmat}{\mat{W}}
\newcommand{\Xmat}{\mat{X}}
\newcommand{\Ymat}{\mat{Y}}
\newcommand{\Zmat}{\mat{Z}}

\newcommand{\xm}[1]{\vec{x}_{m,#1}}

\DeclareMathOperator{\tr}{tr}

\newcommand{\loss}{\ell}

\crefname{equation}{Equation}{Equations}
\crefname{figure}{Figure}{Figures}
\crefname{tabular}{Table}{Tables}
\crefname{subsection}{Section}{Sections}
\crefname{appendix}{Appendix}{Appendices}

\begin{document}

\runningauthor{Pamfil, Sriwattanaworachai, Desai, Pilgerstorfer, Beaumont, Georgatzis, Aragam}

\twocolumn[

\aistatstitle{DYNOTEARS: Structure Learning from Time-Series Data}

\aistatsauthor{Roxana Pamfil\textsuperscript{1$\ast$}, \And Nisara Sriwattanaworachai\textsuperscript{1$\ast$}, \And Shaan Desai\textsuperscript{1$\dagger$}, \And Philip Pilgerstorfer\textsuperscript{1},}
\aistatsauthor{Paul Beaumont\textsuperscript{1}, \And Konstantinos Georgatzis\textsuperscript{1}, \And Bryon Aragam\textsuperscript{2}}
\aistatsaddress{\textsuperscript{1}QuantumBlack, a McKinsey company \And \textsuperscript{2}University of Chicago}
\vspace{-2.2em}
\aistatsaddress{\texttt{\href{mailto:causal@quantumblack.com}{causal@quantumblack.com}} \And \texttt{\href{mailto:bryon@chicagobooth.edu}{bryon@chicagobooth.edu}}}
\vspace{-1.7em}
\aistatsaddress{\textsuperscript{$\ast$}\textit{These authors contributed equally.}}
]

\begin{abstract}
We revisit the structure learning problem for dynamic Bayesian networks and propose a method that simultaneously estimates contemporaneous (\textit{intra-slice}) and time-lagged (\textit{inter-slice}) relationships between variables in a time-series. Our approach is score-based, and revolves around minimizing a penalized loss subject to an acyclicity constraint. To solve this problem, we leverage a recent algebraic result characterizing the acyclicity constraint as a smooth equality constraint. The resulting algorithm, which we call DYNOTEARS, outperforms other methods on simulated data, especially in high-dimensions as the number of variables increases. We also apply this algorithm on real datasets from two different domains, finance and molecular biology, and analyze the resulting output. Compared to state-of-the-art methods for learning dynamic Bayesian networks, our method is both scalable and accurate on real data. The simple formulation and competitive performance of our method make it suitable for a variety of problems where one seeks to learn connections between variables across time.
\end{abstract}

\section{Introduction}

Graphical models are a popular approach to understanding large datasets, and provide convenient, interpretable output that is needed in today's high stakes applications of machine learning and artificial intelligence. In particular, with the growing need for interpretable models and causal insights about an underlying process, directed acyclic graphs (DAGs) have shown promise in many applications. The edges in a DAG provide users with important clues about the relationship between variables in a system. When these edges are not known based on prior knowledge, it is necessary to resort to structure learning, namely, the problem of learning the edges in a graphical model from data. Broadly speaking, structure learning can be divided into static (i.e. equilibrium) and dynamic models, the latter of which explicitly model temporal dependencies.
Static models make sense for independent and identically distributed data.
Many applications, however, exhibit strong temporal fluctuations that we are interested in modeling explicitly.
The problem of learning graphical structures from temporal data collected from dynamic systems has received significant attention from the machine learning \citep{koller2009}, econometrics \citep{lutkepohl2005new}, and neuroscience \citep{rajapakse2007} communities.

In this paper, we revisit the problem of learning \emph{dynamic Bayesian networks} (DBNs) \citep{dean1989model,murphy2002thesis} from data. DBNs have been used successfully in a variety of domains such as clinical disease prognosis \citep{van2008,Zandona:2019aa}, gene regulatory network \citep{linzner2019scalable}, facial and speech recognition \citep{Meng2019,nefian2002}, neuroscience \citep{rajapakse2007}, among others.
DBNs are the standard approach to modeling discrete-time temporal dynamics in directed graphical models.
In econometrics, they are also known as \emph{structural vector autoregressive} (SVAR) models \citep{demiralp2003,swanson1997impulse}.

We propose a simple, score-based approach for learning these models that scales gracefully to high-dimensional datasets. To accomplish this, we cast the problem as an optimization problem (i.e. score-based learning), and use standard second-order optimization schemes to solve the resulting program. Our approach is based on the recent algebraic characterization of acyclicity in directed graphs from \citet{zheng2018}, which makes the formulation simple and amenable to different modeling choices.

\paragraph{Contributions}

The main contributions of this paper are the following: %
\begin{enumerate}[itemsep=-2pt,before=\vspace{-1em},after=\vspace{-1em},label={$\bullet$}]
\item We develop a score-based approach to learning DBNs and use standard optimization routines to optimize the resulting program. The resulting method, which we call DYNOTEARS, can be used to learn time series of arbitrary order, without any implicit assumptions on the underlying graph topologies such as bounded in-degree or treewidth.
\item We validate our approach with extensive simulation experiments, exhibiting the accuracy of our approach in learning both intra-slice and inter-slice relationships in dynamic models.
\item We apply our method to two real datasets: A financial dataset consisting of daily stock returns ($d=97$) and the DREAM4 dataset ($d=100$) \citep{marbach2009generating}. These examples illustrate the importance of modeling both temporal trends as well as steady state relationships and achieve competitive accuracy among other DBN methods.%
\end{enumerate}

The resulting method simultaneously achieves three important goals: 1) Accuracy on high-dimensional data with $d>n$, which allows for application to real world data 2) Robustness to complex graph topologies, and 3) A simple, plug-n-play algorithm for learning based on black-box optimization.

\paragraph{Related work}

There are many methods for learning DBNs in the literature. Some approaches ignore contemporaneous dependencies and recover only time-lagged relationships \citep{haufe2010sparse,song2009}. Others learn both types of relationships independently \citep{haufe2010sparse,song2009}. Many methods follow a two-step approach of first learning inter-slice weights and then estimating intra-slice weights from the residuals from the first step \citep{chen2007,hyvarinen2010,moneta2011causal}. There are also hybrid algorithms that combine conditional-independence tests and local search to improve the score \citep{malinsky2018,malinsky2019}. While all of these methods can achieve good structure recovery on small graphs, they suffer from the curse of dimensionality. More discussion and comparison can be found in \cref{subsec:alternativeFormulations}.
There is also an extensive literature on learning SVAR models in the econometrics and statistics literature \citep{demiralp2003,lanne2017identification,reale2001identification,reale2002sampling,swanson1997impulse,tank2019identifiability}.

Since algorithms for learning DBNs typically rely internally on calling methods for learning static BNs, it is worth briefly reviewing this here. These methods can be classified into
\textit{constraint-based} methods and \textit{score-based} methods, as well as hybrid methods that combine these two approaches. Constraint-based methods use conditional independence tests to recover the Markov equivalence class of DAGs under the assumption of faithfulness \citep[e.g.][]{colombo2012,spirtes1991}. While this yields fast algorithms,
these methods are sensitive to the underlying graph structure and suffer from error propagation \citep{spirtes2010introduction}. Score-based methods, on the other hand, use a score function to find the best DAG that fits the given data \citep[e.g.][]{Heckerman95,chickering2002,Bouckaert93prob}. %
Examples of scores include the BIC, BDe, and BDeu \citep{spirtes2000}. Score-based methods are computationally expensive due to the acyclicity constraint and the vast number of DAGs to search over \citep{robinson1977}.
The recent work of \citet{zheng2018} expresses the acyclicity of a DAG by a smooth equality constraint,
which makes it possible to formulate structure learning as a smooth minimization problem subject to this equality constraint.

\section{Dynamic Structure Learning}

\begin{figure}[t]
	\centering
	\includegraphics[width=\columnwidth]{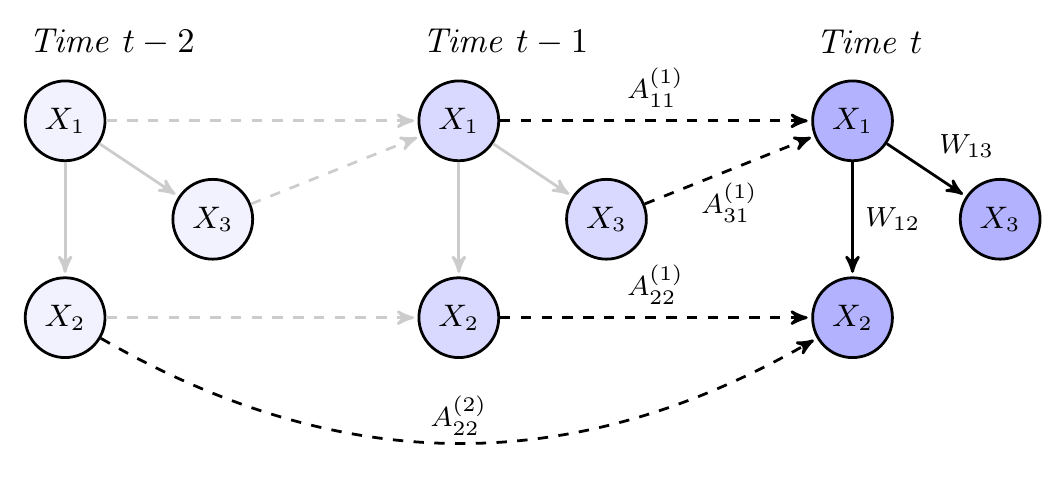}
	\caption{Illustration of \textit{intra-slice} (solid lines) and \textit{inter-slice} (dashed lines) dependencies in a DBN with $d=3$ nodes and autoregression order $p=2$. For clarity, we display edges that do not influence the variables at time $t$ in a lighter shade.}
	\label{fig:diagram}
\end{figure}

\subsection{Formulation}

Consider $M$ independent realizations of a stationary time series, with the $m\mathrm{th}$ time series given by $\{\vec{x}_{m,t}\}_{t \in \{0,\ldots,T\}}$ for $\vec{x}_{m,t}\in\mathbb{R}^d$, where $d$ represents the number of variables in the dataset. We assume that variables influence each other in both a contemporaneous and a time-lagged manner, as illustrated in \cref{fig:diagram}. We call these \textit{intra-slice} and \textit{inter-slice} dependencies, respectively.

We model the data using the following standard SVAR model \citep{demiralp2003,kilian2011structural,swanson1997impulse}:
\begin{equation}\label{eqn:dynSEMvec}
\xm{t}^\top = \xm{t}^\top\Wmat + \xm{t-1}^\top\Amat_1 + \ldots + \xm{t-p}^\top\Amat_p + \vec{z}_{m,t}^\top \
\end{equation}
for $t \in \{p,\ldots,T\}$ and for all $m \in \{1,\ldots,M\}$, where $p$ is the \textit{autoregressive order}, and $\vec{z}_{m,t}$ is a vector of centered error variables, which are independent within and across time. The error variables are \emph{not} assumed to be Gaussian. Extensions to nonlinear models are also possible, but not explored here, see Section~\ref{sec:disc} for more discussion.

The matrices $\Wmat$ and $\Amat_i$ ($i \in \{1,\ldots,p\}$) represent weighted adjacency matrices with nonzero entries corresponding to the intra-slice and inter-slice edges, respectively. When $\Wmat$ is acyclic, as we will assume throughout this paper, these matrices parametrize a DBN. We also assume that this network structure is constant across time. %
This allows us to write
\cref{eqn:dynSEMvec}
in matrix form as
\begin{equation}\label{eqn:dynSEM}
\Xmat = \Xmat \Wmat + \Ymat_1\Amat_1 + \ldots + \Ymat_p\Amat_p + \Zmat,
\end{equation}
where $\Xmat$ is an $n \times d$ matrix whose rows are $\xm{t}^\top$, and the matrices $\Ymat_1,\ldots,\Ymat_p$ are time-lagged versions of $\Xmat$. The number $n$ of rows is our effective sample size, and we have $n=M(T+1-p)$.

Let $\Ymat=[~\Ymat_{1} ~|~ \cdots ~|~ \Ymat_{p}~]$ be the $n \times pd$ matrix of time-lagged data. Additionally, let $\Amat=[~\Amat_1^\top ~|~ \cdots ~|~ \Amat_p^\top~]^\top$ be the $pd \times d$ matrix of inter-slice weights. With this notation, the SEM in \labelcref{eqn:dynSEM}  takes the compact form:
\begin{equation}\label{eqn:dynSEMcompact}
\Xmat = \Xmat \Wmat + \Ymat\Amat + \Zmat \,.
\end{equation}

This general formulation makes it possible to consider scenarios in which the time-lagged data matrix $\Ymat$ does not necessarily cover a contiguous sequence of time slices (i.e., from $t-p$ to $t-1$). For instance, in time series that exhibit known seasonality patterns, one can include in the lagged data matrix $\Ymat$ only those time points that have an impact on the variables at time $t$.

\paragraph{Identifiability}
Identifiability in SVAR models is a central topic of the econometrics literature (see \citealp{kilian2011structural} for a review). Identifiability of $\Amat$ follows from standard results on vector autoregressive (VAR) models, whereas identifiability of $\Wmat$ is more difficult to establish. We focus on two special cases where identifiability holds:
\begin{enumerate}[itemsep=-2pt,before=\vspace{-1em},after=\vspace{-1em},label={$\bullet$}]
\item The errors $\vec{z}_{m,t}$ are non-Gaussian. Identifiability in this model is a well-known consequence of Marcinkiewicz's theorem on the cumulants of the normal distribution \citep{kagan1973characterization,marcinkiewicz1939propriete} and independent component analysis (ICA), see \citet{hyvarinen2010,lanne2017identification,moneta2011causal}.
\item The errors $\vec{z}_{m,t}$ are standard Gaussian, i.e. $\vec{z}_{m,t}\sim\mathcal{N}(0,I)$. Identifiability of this model is an immediate consequence of Theorem~1 of \citet{peters2013identifiability} and the acyclicity of $\Wmat$.
\end{enumerate}

\noindent
In what follows, we assume that one of these two conditions on $\vec{z}_{m,t}$ holds.

\subsection{Optimization problem}\label{subsec:optimizationProblem}

Given the data $\Xmat$ and $\Ymat$,
our goal is to estimate weighted adjacency matrices $\Wmat$ and $\Amat$ that correspond to DAGs. The edges in $\Amat$ go only forward in time and thus they do not create cycles.
In order to ensure that the whole network is acyclic, %
it thus suffices to require that $\Wmat$ is acyclic. Minimizing the least-squares loss with the acyclicity constraint gives the following optimization problem:
\begin{gather}
\min_{\Wmat,\,\Amat } ~~ \loss(\Wmat,\Amat)
~~ \mbox{s.t.} ~~ \text{$\Wmat$ is acyclic}\,,\\
~~ \textup{where}\,\, \loss(\Wmat,\Amat)
= \frac{1}{2n}\lVert \Xmat - \Xmat \Wmat - \Ymat\Amat \rVert_F^2 \notag.
\end{gather}
To enforce the sparsity of $\Wmat$ and $\Amat$, we also introduce $\ell_1$ penalties in the objective function. Let the regularized optimization problem be
\begin{gather}\label{eqn:l1optimizationProblem}
\min_{\Wmat,\,\Amat } ~~ f(\Wmat,\Amat)%
~~ \mbox{s.t.} ~~ \text{$\Wmat$ is acyclic}\,,\\
~~ \textup{with}\,\, f(\Wmat, \Amat) = \loss(\Wmat,\Amat)
+ \lambda_{\Wmat}\lVert \Wmat \rVert_1 + \lambda_{\Amat}\lVert \Amat \rVert_1\,,\notag
\end{gather}
where $\lVert \cdot \rVert_1$ stands for the element-wise $\ell_1$ norm.
Regularization is especially useful in cases with much fewer samples than variables, $n \ll d$.
For example, for static BNs, the least-squares loss has been shown to be consistent in high-dimensional settings for learning DAGs \citep{van2013ell_,aragam2015learning}.

The key difficulty in solving the optimization problem \eqref{eqn:l1optimizationProblem} is the acyclicity constraint on $\Wmat$. To deal with this, we use an equivalent formulation of acyclicity via the trace exponential function, due to \citet{zheng2018}. They show that the function $h(\Wmat)=\tr e^{\Wmat\circ\Wmat}-d$ satisfies $h(\Wmat)=0$ if and only if $\Wmat$ is acyclic. Here, $\circ$ denotes the Hadamard product of two matrices. Replacing the acyclicity constraint in \eqref{eqn:l1optimizationProblem} with the equality constraint $h(\Wmat)=0$, the resulting equality constrained program can be solved as in \citet{zheng2018}, using the augmented Lagrangian method. This translates the problem to a series of unconstrained problems of the form
\begin{equation}\label{eqn:FullOptimizationProblem}
\min_{\Wmat,\,\Amat }{F(\Wmat,\Amat)}\, %
\end{equation}
where $F$ is the following smooth augmented objective:
\begin{equation}\label{eqn:optimUnconstrained}
F(\Wmat, \Amat) = f(\Wmat,\Amat) + \frac{\rho}{2}h(\Wmat)^2+\alpha h(\Wmat)\,.
\end{equation}

By writing $\Wmat=\Wmat_+-\Wmat_-$ such that $\Wmat_+\geq0$ and $\Wmat_-\geq0$ (and analogously for $\Amat$), we transform \labelcref{eqn:FullOptimizationProblem} to a quadratic program with non-negative constraints and twice the number of variables. %
The resulting problem can be solved using standard solvers such as \texttt{L-BFGS-B} \citep{zhu1997algorithm}.

Following \citet{zheng2018}, in order to reduce the effect of numerical error from computing $h(\Wmat)$, we eliminate weights close to 0 by thresholding the entries of $\Wmat$ and $\Amat$. %
Overall, DYNOTEARS has $4$ hyperparameters: the regularization constants $\lambda_{\Wmat}$ and $\lambda_{\Amat}$, and the weight thresholds $\tau_{\Wmat}$ and $\tau_{\Amat}$.

\subsection{Alternative formulations}\label{subsec:alternativeFormulations}

An alternative approach to jointly minimizing \eqref{eqn:FullOptimizationProblem} over $(\Wmat,\Amat)$ is to use a two-stage optimization procedure, similar to those in \citet{chen2007,demiralp2003,hyvarinen2010}. We can rewrite \Cref{eqn:dynSEM} as the structural VAR (SVAR)
\begin{equation}\label{eqn:SVAR}
\Xmat\Amat_0 = \Ymat_1\Amat_1+\ldots+\Ymat_{p}\Amat_p+ \Zmat\,,
\end{equation}
where $\Amat_0=\Imat-\Wmat$. With the acyclicity constraint, $\Wmat$ can be written as an upper triangular matrix. It follows that $\Amat_0$ has full rank and is invertible. Right-multiplying \labelcref{eqn:SVAR} by $\Amat_0^{-1}$ gives
\begin{equation}\label{eqn:transformedVAR}
\Xmat=\Ymat_{1}\Bmat_1 + \ldots + \Ymat_{p}\Bmat_p + \emat\,,
\end{equation}
where $\Bmat_i=\Amat_i\Amat_0^{-1}$ ($i \in \{1,\ldots,p\}$) and $\emat=\Zmat\Amat_0^{-1}$ is a noise term whose elements are correlated (for
fixed $t$). %
\Cref{eqn:transformedVAR} is now a \emph{reduced-form} VAR that we can fit using least-squares \citep{amisano2012topics}. This produces estimates of the matrices $\Bmat_i$ and of the residuals $\emat$. Recall that
\begin{equation}
\emat\Amat_0=\Zmat \Leftrightarrow \emat(\Imat-\Wmat) = \Zmat \Leftrightarrow \emat = \emat\Wmat + \Zmat \,,
\end{equation}
so we can estimate $\Wmat$ by applying static NOTEARS to the matrix of residuals $\emat$.

One disadvantage of the two-step approach is that enforcing sparsity via $\ell_1$ regularization is no longer straightforward, as $\Bmat_i$ and $\Amat_i$ do not necessarily have the same sparsity patterns. Moreover, any errors in estimating the $\Bmat_i$ at the right sparsity levels propagate to the residuals $\emat$, which directly affects the estimation of $\Wmat$. In turn, this affects the final step of the process, where one calculates $\Amat_i=\Bmat_i(\Imat-\Wmat)$. See \cref{sec:oneStageVsTwoStage} for additional discussion. %

\section{Experiments}\label{sec:numericalExperiments}

To validate the effectiveness of the proposed method, we performed a series of simulation experiments for which the ground truth is known in advance. We focus here on the main results; more detailed results can be found in \Cref{subsec:numericalExperiments}. %

\subsection{Setup}

To benchmark DYNOTEARS against existing approaches, we simulate data according to the SEM from \labelcref{eqn:dynSEMcompact}. There are three steps to this process: 1) generating the weighted graphs $\mathcal{G}_{\Wmat}$ and $\mathcal{G}_{\Amat}$, 2) generating data matrices $\Xmat$ and $\Ymat$ consistent with these graphs, and 3) running all algorithms on $\Xmat$ and $\Ymat$ and computing performance metrics. See \Cref{subsec:dataGenProcess} for a detailed description of steps 1) and 2).

There are three algorithms that we use for benchmarking. The first algorithm is based on a general approach from \citet{murphy2002thesis}; here, we use static NOTEARS and Lasso regression to estimate $\Wmat$ and $\Amat$ independently. The second algorithm is the SVAR estimation method based on LiNGAM \citep{hyvarinen2010}. The third algorithm is tsGFCI \citep{malinsky2018}. For details, see \cref{subsec:otherAlgorithms}.

\begin{figure*}[ht!]
	\centering
	\includegraphics[width=0.8\textwidth]{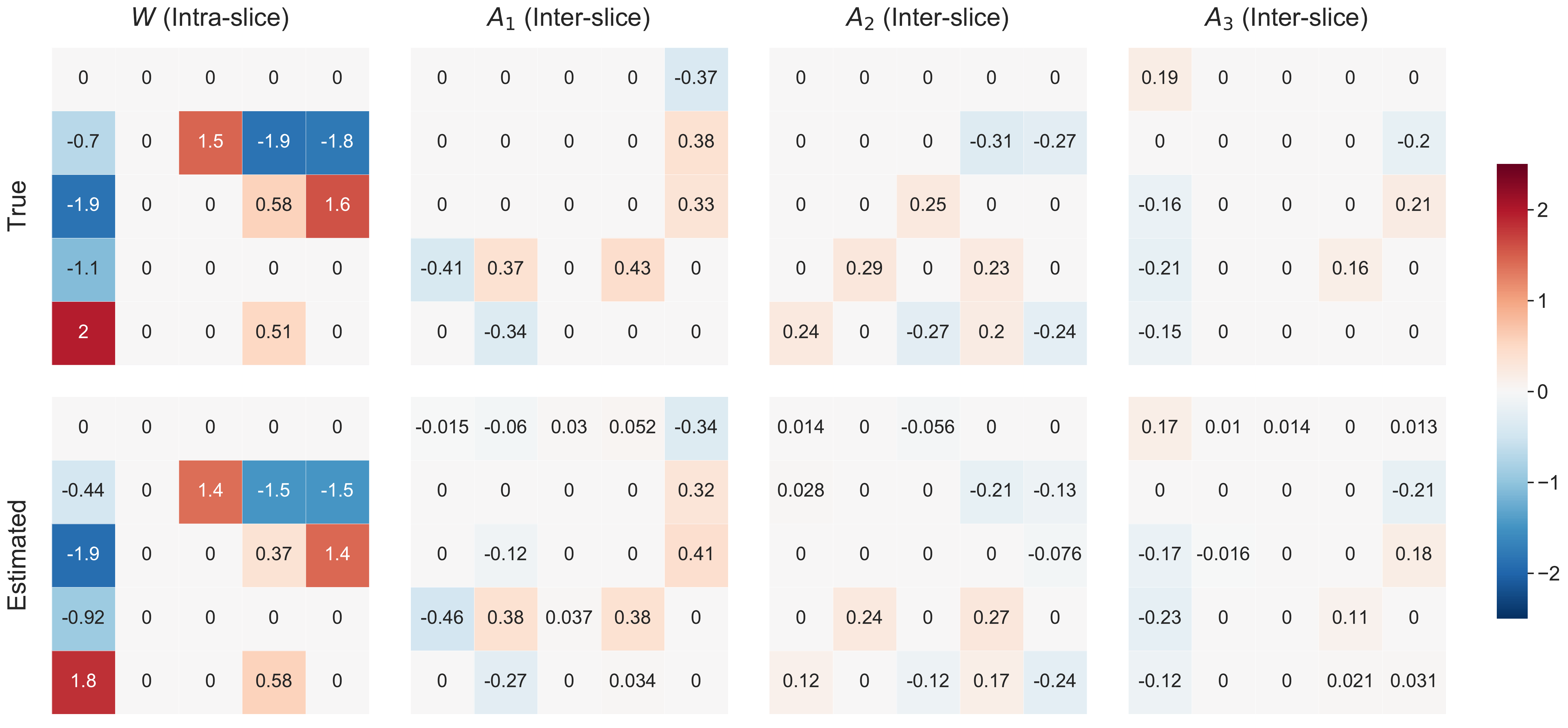}
	\caption{Example results using DYNOTEARS for data with Gaussian noise, $n=500$ samples, $d=5$ variables, and $p=3$ autoregressive terms. We set the regularization parameters $\lambda_{\Wmat}=\lambda_{\Amat}=0.05$ and the weight thresholds $\tau_{\Wmat}=\tau_{\Amat}=0.01$. Our algorithm recovers weights that are close to the ground truth.}
	\label{fig:exampleOutputp3}
\end{figure*}

\subsection{Results}\label{subsec:simulationResults}

\begin{figure*}[t!]
    \centering
    \begin{subfigure}{0.78\columnwidth}
        \includegraphics[width=\textwidth]{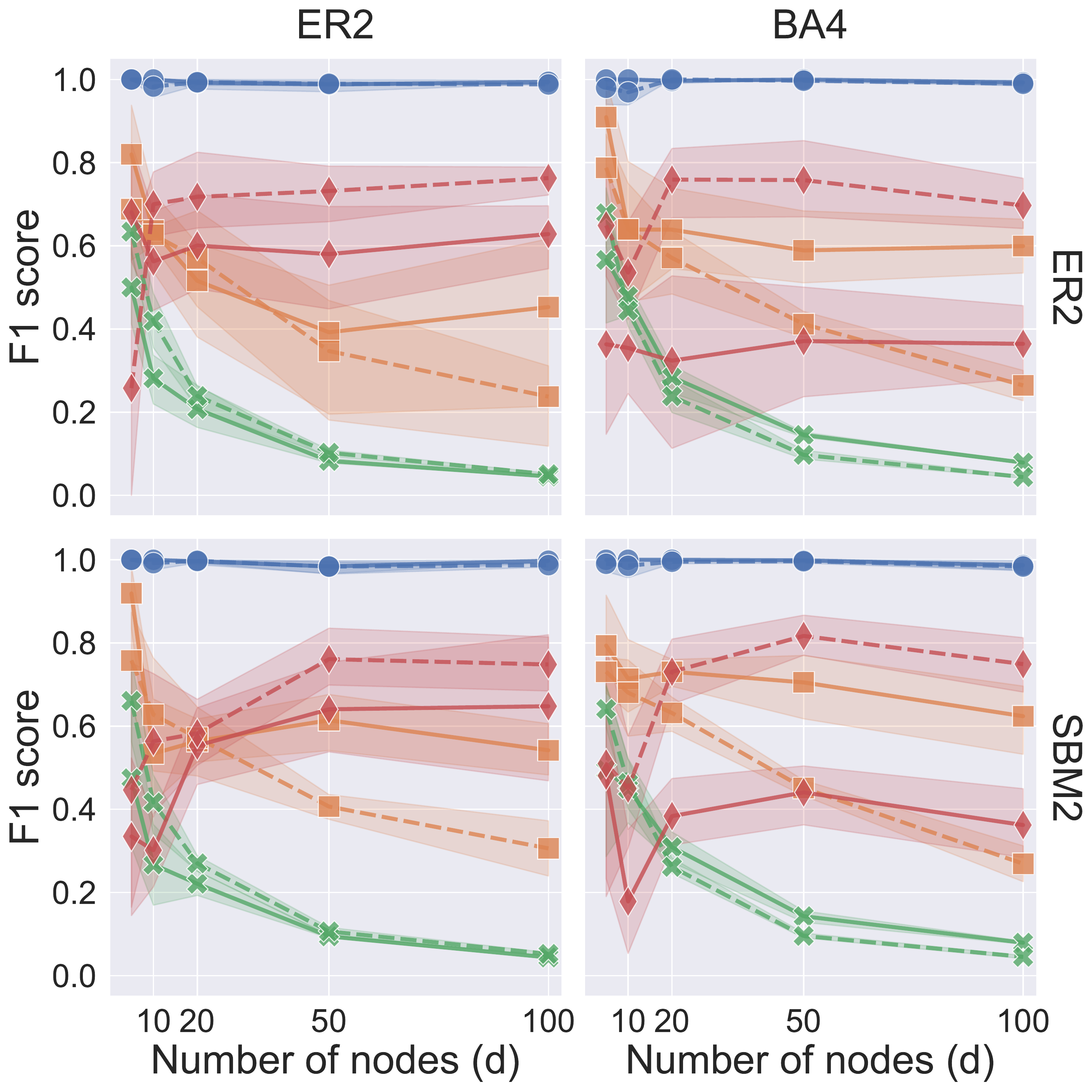}
        \caption{Gaussian noise, $n=500$}
        \label{fig:sim_Gauss_n500}
    \end{subfigure} \hspace{2em}
    \begin{subfigure}{0.78\columnwidth}
        \includegraphics[width=\textwidth]{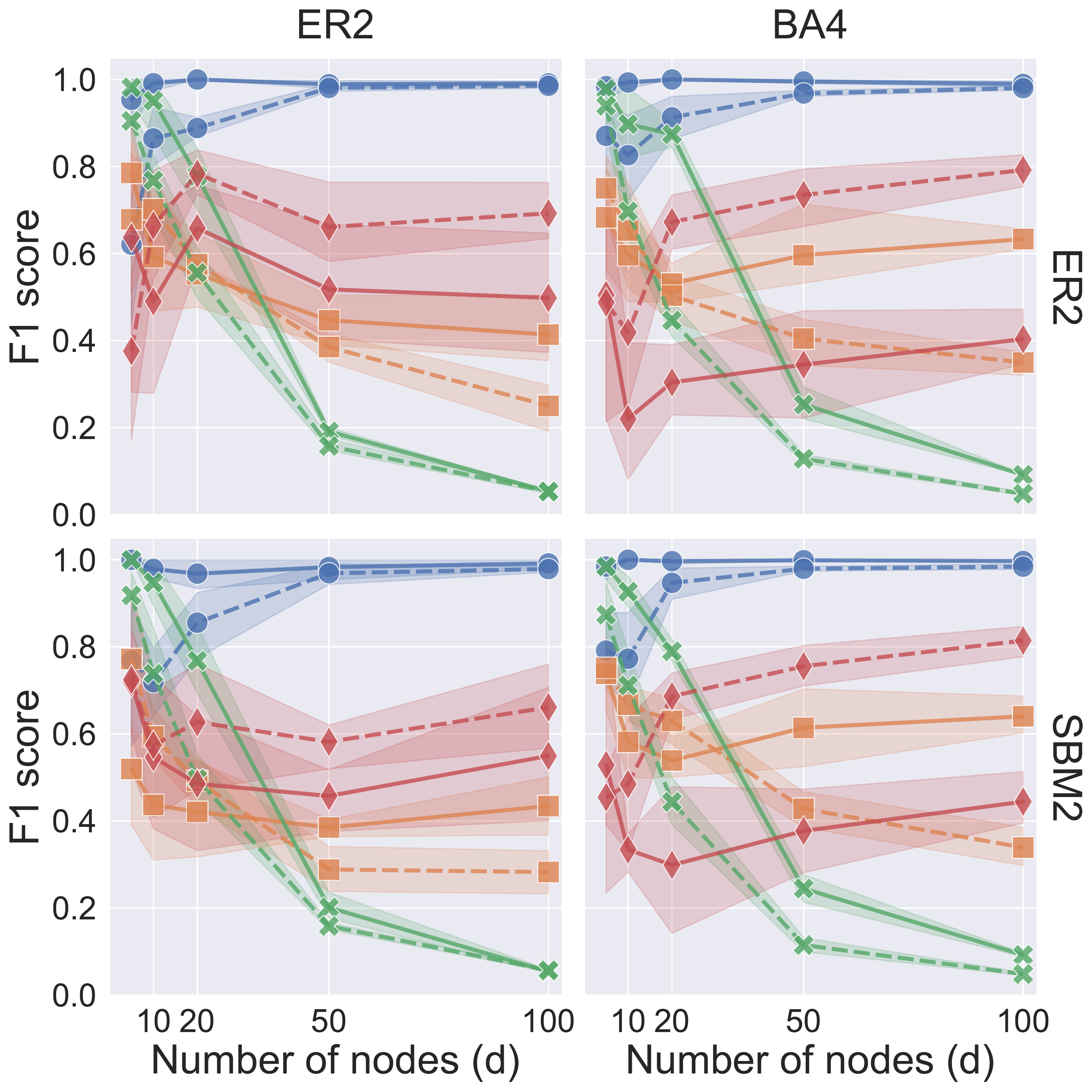}
        \caption{Exponential noise, $n=500$}
        \label{fig:sim_exp_n500}
    \end{subfigure} \\
    \begin{subfigure}{0.78\columnwidth}
        \includegraphics[width=\textwidth]{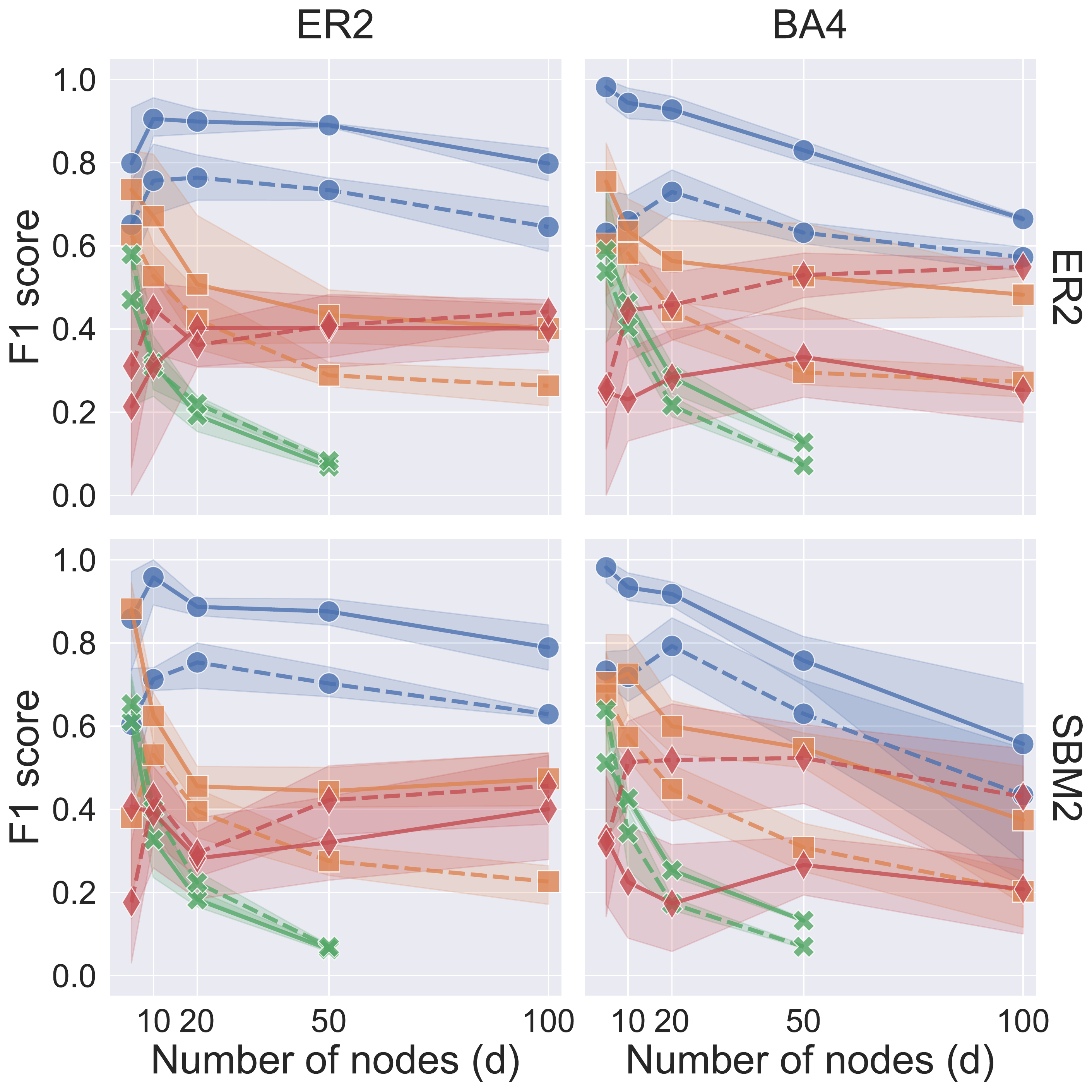}
        \caption{Gaussian noise, $n=50$}
        \label{fig:sim_Gauss_n50}
    \end{subfigure} \hspace{2em}
    \begin{subfigure}{0.78\columnwidth}
        \includegraphics[width=\textwidth]{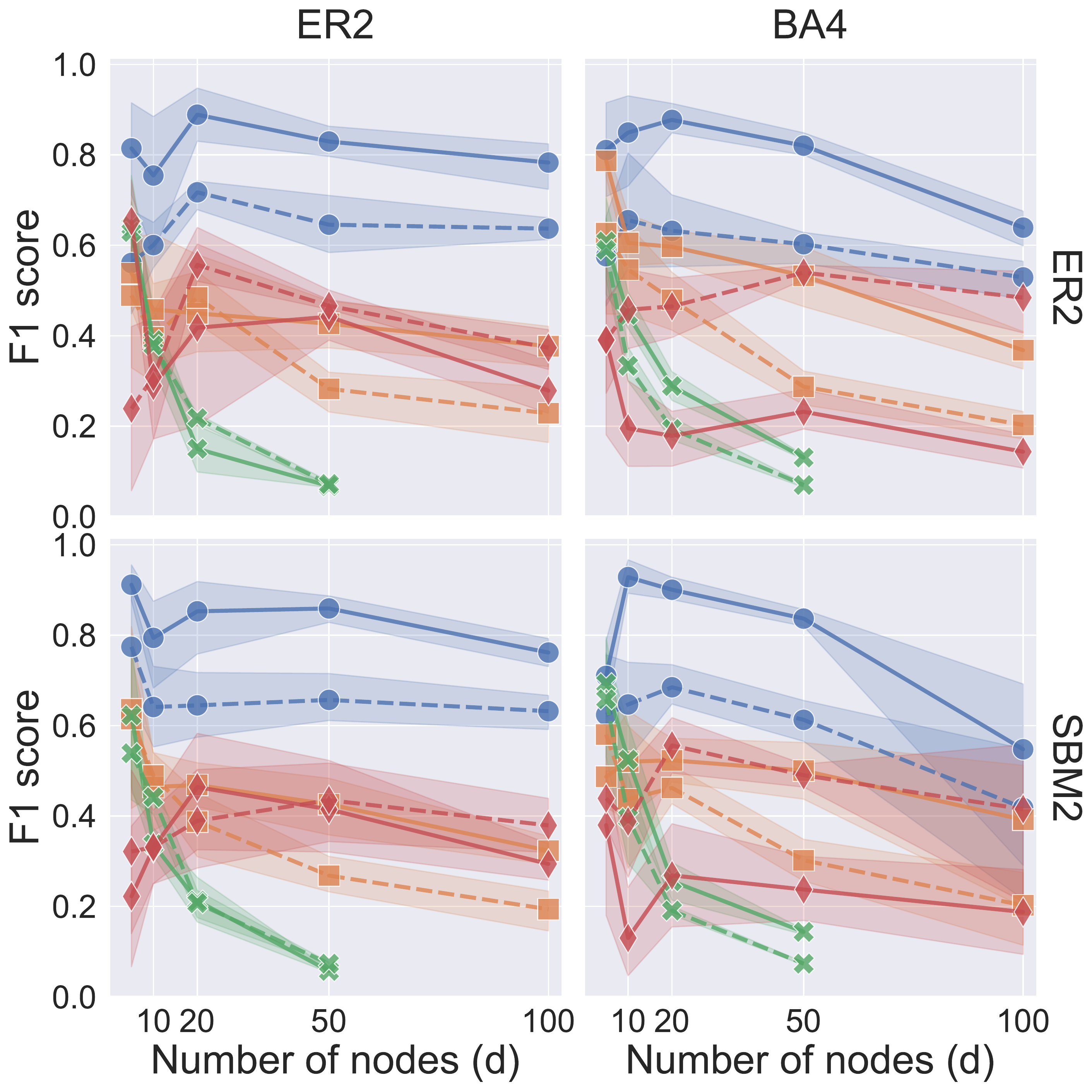}
        \caption{Exponential noise, $n=50$}
        \label{fig:sim_exp_n50}
    \end{subfigure} \\
    \includegraphics[width=0.8\textwidth]{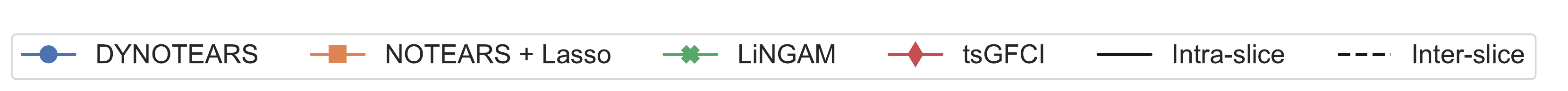}
    \caption{F1 scores for different noise models (Gaussian, Exponential) and different sample sizes ($n \in \{50,500\}$.). Each panel contains results for two different choices of intra-slice graphs (columns) and inter-slice graphs (rows). Every marker corresponds to the mean performance across $5$ algorithm runs, each on a different simulated dataset. F1 scores for intra-slice and inter-slice edges appear in continuous and dashed lines, respectively. DYNOTEARS typically outperforms the other algorithms. Note that the ICA step in LiNGAM does not work for $n<d$, so the corresponding markers for $d=100$ are missing from panels (c) and (d).}
    \label{fig:simulationsF1}
\end{figure*}

We start by illustrating the typical performance of DYNOTEARS on one simulated dataset in \cref{fig:exampleOutputp3}. We follow the process from \Cref{subsec:dataGenProcess} to generate data with Gaussian noise, $n=500$ samples, $d=5$ variables, and $p=3$ autoregressive terms. The intra-slice DAG is an ER graph with mean degree equal to $4$ (counting edges in either direction), and the inter-slice DAG is an ER graph with a mean out-degree equal to $1$. %
The base of the exponential decay of inter-slice weights is $\eta=1.5$. We run DYNOTEARS on this dataset with regularization parameters $\lambda_{\Wmat}=\lambda_{\Amat}=0.05$ and weight thresholds $\tau_{\Wmat}=\tau_{\Amat}=0.01$. The low values for the thresholds are intentional, as we want to highlight that the algorithm recovers a structure close to the ground truth even without tweaking these parameters. We consider here the case where the autoregressive order $p$ is known, however, see \cref{subsec:autoregressiveOrder} for details on how to estimate $p$ when it is unknown. %
As \cref{fig:exampleOutputp3} shows, estimated weights are close to the true weights for both $\Wmat$ and $\Amat$. The only significant omission is a missing entry in the third row and third column of $\Amat_2$. Performance is similar without $\ell_1$ regularization and for larger values of $p$.

In \cref{fig:simulationsF1}, we compare the performance of DYNOTEARS to that of three other algorithms (see \cref{subsec:otherAlgorithms}). We consider two distributions for the noise, Gaussian and Exponential, and two choices for the number of samples, $n \in \{50,500\}$. Within each of the four panels, each column corresponds to one choice of intra-slice graph model and mean degree; for instance, ER2 indicates that we used an Erd\H{o}s--R\'{e}nyi graph model with a mean degree of 2. Similarly, each row corresponds to one choice of inter-slice graph model and mean degree. For each individual plot, we generate $n$ data samples with autoregression order $p=1$ for five choices of the number of variables, $d \in \{5,10,20,50,100\}$. The vertical axis indicates the performance of each algorithm in terms of the F1 score, which we calculate separately for intra-slice and inter-slice matrices. We discuss the selection of hyperparameter values for the four algorithms in \cref{subsec:hyperparameters}. The relative ranking of the four algorithms is not especially sensitive to these hyperparameters. In particular, the regularization parameters are largely irrelevant when there is sufficient data ($n \gg dp$).

DYNOTEARS is the best-performing algorithm in \cref{fig:simulationsF1}, with F1 scores close to 1 for $n=500$. DYNOTEARS is also the best-performing algorithm when the number of variables exceeds the number of samples (see panels (c) and (d) for $d=100$). This high-dimensional case is especially difficult, yet common in applications, and we discuss one such example in \cref{subsec:DREAM}. The second-best algorithm is tsGFCI. However, its performance tends to degrade as we add more edges to the ground-truth graphs; see \cref{fig:simGaussF1} in the Appendix. The variance in performance is also larger for tsGFCI than for the other algorithms.
As expected, learning intra-slice and inter-slice structure separately with NOTEARS + Lasso underperforms DYNOTEARS. In particular, we find that the Lasso step falsely identifies some intra-slice edges as inter-slice. LiNGAM is an algorithm designed for non-Gaussian data, so its poor performance in panels (a) and (c) of \cref{fig:simulationsF1} is not surprising. However, even on data with exponential noise (panels (b) and (d) of \cref{fig:simulationsF1}), its performance degrades significantly as $d$ increases. \cref{subsec:additionalResults} contains additional figures that compare the performance of the four algorithms using metrics other than the F1 score.

\section{Applications}\label{sec:applications}

Our approach allows us to detect whether contemporaneous or time-lagged relationships are more meaningful in a given dataset.
We discuss two examples for which each type of interaction is dominant.

\subsection{S\&P 100 stock returns}

We apply DYNOTEARS to a financial dataset of daily stock returns of companies in the S\&P 100. The dataset was obtained using the \texttt{yahoofinancials}\footnote{\url{https://pypi.org/project/yahoofinancials/}} Python package and consists of daily closing prices from 1 July 2014 to 30 June 2019, inclusive. To account for non-stationarity in the stock price data, we use log-returns, defined as the temporal difference of the logarithms of the stock prices.
We test the stationarity of the transformation using the Augmented Dick-Fuller test and reject the Null hypothesis of a unit-root with $p_\mathrm{value}<0.01$.
We normalise the log-returns so that they have zero mean and unit variance. If we did not perform this step, the regularization would favor low-variance stocks, whose (linear) weights would be relatively larger than for high-risk stocks.

The resulting dataset contains $n=1257$ samples (i.e., trading days) and $d=97$ variables (i.e., stocks).
We apply DYNOTEARS with $p=1$, $\lambda_{\Wmat}=0.1$, and $\lambda_{\Amat}=0.1$.
The values are selected via grid search; see \cref{fig:validation_sp100} in the supplement. We hold out the final 400 data points of the series for validation ($\approx32\%$) and we discard 2 trading days to avoid validation-set contamination.
Running DYNOTEARS on the whole dataset took $3.8$ minutes on a laptop, with $25.8$ minutes total CPU time. %

The final graph does not contain inter-slice edges. This means that the best prediction of future returns is the current return. This could be evidence for the efficient market hypothesis, which says that asset prices contain all information available \citep{malkiel1970efficient}.
While there are no inter-slice edges, the validation loss for DYNOTEARS is slightly smaller than for static NOTEARS with the same $\lambda_{\Wmat}$ value, $167.387 < 167.784$. Comparing the two graphs, the biggest difference is that the edge between \texttt{GOOGL} and \texttt{GOOG} flips.\footnote{Both shares are issued by Alphabet, the holding company of Google, and they are equal in terms of dividends. However, only \texttt{GOOGL} includes shareholder voting rights.} The dynamic parametrization helps to find a better local optimum.

\begin{figure*}[th!]
	\centering
	\includegraphics[width=1.3\columnwidth]{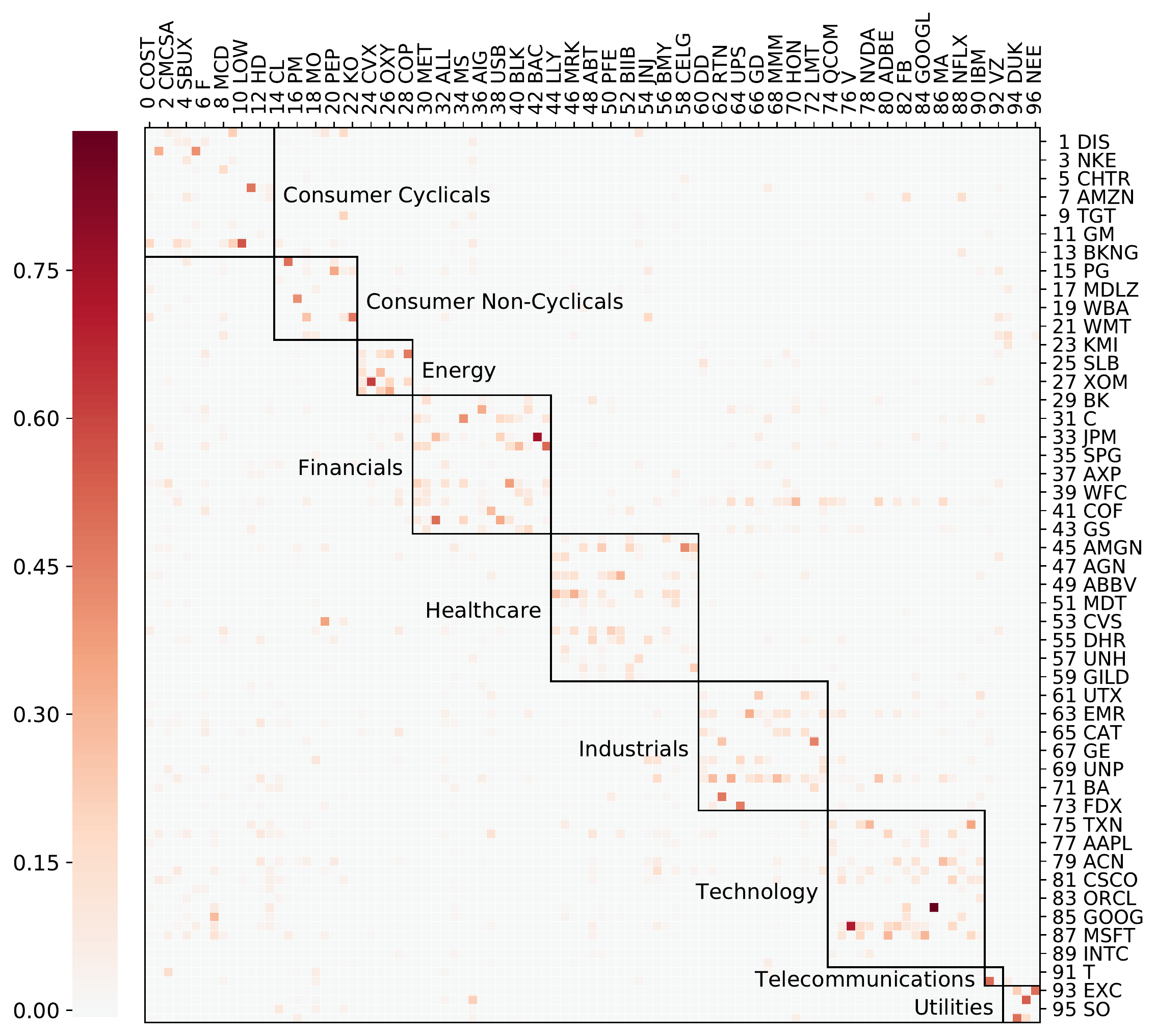}
	\caption{Estimated intra-slice matrix $\Wmat$, with rows and columns sorted by sector. A row is the source stock, a column is the target stock. Stocks in the same sector tend to have larger edge weights between them. The normalized mutual information between a community-induced partition and the sector-based partition indicates agreement between the estimated network and the sectors.}
	\label{fig:stocksW}
\end{figure*}

\Cref{fig:stocksW} provides a visualization of the estimated intra-slice weights matrix $\Wmat$.
When we order the stocks by industry sector, we obtain an approximately block-diagonal structure.
Thus, two stocks are more likely to influence each other if they belong to the same sector than if they are from different sectors.
The weight of $96\%$ of all identified edges is positive, so most stocks move together. This percentage is roughly the same for edges within and between sectors. One noteworthy feature in \cref{fig:stocksW} is that Amazon (\texttt{AMZN}), which is part of the Consumer Cyclicals sector, is connected to many of the technology stocks, including Facebook, Netflix, NVIDIA, Google, and Microsoft.

To quantify the sector relationship, we apply the Louvain method \citep{Blondel_2008}, a community detection algorithm and a form of graph clustering, to $\Wmat$.\footnote{We use the \emph{python-louvain} package from \url{https://pypi.org/project/python-louvain/}.} This partitions the set of stocks into subsets that correspond to densely-connected subgraphs in the estimated Bayesian network. The normalized mutual information (NMI, \citealp{li2001}) between this community-induced partition and the sector-based partition is approximately $0.79$, which indicates close agreement.\footnote{An NMI value of $1$ corresponds to identical partitions, whereas a value of $0$ corresponds to independent partitions.}

\subsection{DREAM4 gene expression data}\label{subsec:DREAM}

We benchmark DYNOTEARS on the DREAM4 network inference challenge \citep{marbach2009generating}, in which the objective is to learn gene regulatory networks from gene expression data. DREAM4 consists of 5 independent datasets, each with 10 different time-series recordings for 100 genes across 21 time steps. We preprocess our data and choose hyperparameters $\lambda_{\Amat}$ and $\lambda_{\Wmat}$ through 10-fold cross validation, details of which can be found in \cref{sec:cvDream}.
In \citet{lu2019causal}, the authors compared different approaches to learning these networks, including methods based on mutual information, Granger causality, dynamical systems, decision trees, Gaussian processes (GPs), and DBNs.
Unsurprisingly, their results indicate that flexible nonparametric methods such as GPs and decision trees perform the best. It is nonetheless instructive to compare the performance of DYNOTEARS to these methods for two reasons: 1) This gives us a sense of how much is lost in assuming the linear model \eqref{eqn:dynSEMvec}, and 2) For an apples-to-apples comparison, we can still compare DYNOTEARS to the DBN methods (six in total) tested by \citet{lu2019causal}.

\begin{table}[ht!]
	\centering
	\begin{tabular}{l|c|c}
		\toprule
	  	Algorithm &  Mean AUPR & Mean AUROC \\
		\midrule
	    DYNOTEARS &         \textbf{0.173} &         0.664 \\
			G1DBN &         0.110 &         \textbf{0.676}\\
			ScanBMA &       0.101 &         0.657\\
			VBSSMb &        0.096 &         0.618\\
			VBSSMa &        0.086 &         0.624\\
	    Ebdbnet &       0.043 &         0.643\\
		\bottomrule
	\end{tabular}
  \caption{AUPR and AUROC scores of DBN algorithms on the DREAM4 dataset. Values for methods other than DYNOTEARS are from \citet{lu2019causal}.}
  \label{table:auprSmall}
\end{table}

As in the original DREAM4 challenge, we use mean AUPR and AUROC across the 5 datasets to compare DYNOTEARS to the 23 algorithms presented in \citet{lu2019causal}. The results are as follows:
\begin{enumerate}[itemsep=-2pt,before=\vspace{-1em},after=\vspace{-1em},label={$\bullet$}]
\item DYNOTEARS achieves an average AUROC of 0.664 and an average AUPR of 0.173. Among the six DBN methods tested, this ranks \textbf{1st} and \textbf{2nd} in AUPR and AUROC, respectively, as shown in \Cref{table:auprSmall}.
\item Furthermore, DYNOTEARS is within one standard deviation of the best performing method (G1DBN) based on AUPR, and no other method is within one standard deviation of DYNOTEARS based on AUROC.
\item Overall, this ranks \textbf{4th} in AUPR and \textbf{8th} in AUROC (see \Cref{table:aupr} and \Cref{table:auroc}, respectively). While the top performing methods were based on nonparametric models such as GPs, DYNOTEARS still outperforms several other nonparametric methods despite its use of the linear model \eqref{eqn:dynSEMvec}.
\end{enumerate}

\section{Discussion}
\label{sec:disc}

In this paper, we proposed DYNOTEARS, an algorithm for learning dynamic Bayesian networks, inspired by recent work on structure learning for static Bayesian networks using differentiable acyclicity constraints \citep{zheng2018}.
Our algorithm learns both intra-slice and inter-slice dependencies between variables simultaneously, in contrast with some existing methods that perform these estimations in succession.

An important feature of DYNOTEARS is its simplicity, both in terms of formulating an objective function and in terms of optimizing it.
It performs well on simulated data across a wide range of parameter choices in the data-generation process. %

We also applied DYNOTEARS to two empirical datasets from different application domains, finance and molecular biology. The results reveal insightful patterns in the data. %
Both of these applications have $d \approx 100$, confirming that DYNOTEARS can be applied to larger datasets than those considered in most existing work on DBNs.

To conclude, we briefly discuss some limitations and possible extensions of DYNOTEARS.

\paragraph{Assumptions}
We have assumed that the structure of the DBN is fixed through time and is identical for all time series in the data (i.e., it is the same for all $m \in \{1,\ldots, M\}$). It would be useful to relax these assumptions in various ways, for example by allowing the structure to change smoothly over time \citep{song2009} or at discrete change points that we infer from the data \citep{grzegorczyk2011}.
Another topic for future work is to investigate the behaviour of the algorithm on nonstationary or cointegrated time series \citep{malinsky2019}, or in situations with confounders \citep{huang2015,malinsky2018}. A possible approach is to apply a post-processing step to the output of DYNOTEARS so as to remove spurious relationships between variables (e.g., by using statistical tests).

\paragraph{Undersampling}
As pointed out by a reviewer, as with most DBN models, we implicitly assume that the sampling rate of the process is at least as high as the fluctuations in the underlying causal process. See for example \citet{gong2015discovering,hyttinen2016causal,plis2015rate,cook2017learning}.
As a check on the sensitivity of DYNOTEARS to this (strong) assumption, we tested a modification of our approach for undersampled data adapted from \citet{cook2017learning}. In essence, by running DYNOTEARS on the data and then post-processing any intra-slice edges by making them bi-directed, we can obtain a graph which is comparable to the methods suggested in \citet{cook2017learning}. Although our method was not designed to handle undersampling, it achieves lower false-negative rates compared to other methods. Of course, more careful modifications to handle undersampling is an interesting direction for future work.

\paragraph{Nonlinear dependence}
Finally, we emphasize that linear assumption in (\ref{eqn:dynSEMvec}-\ref{eqn:dynSEM}) is made purely for simplicity, in order to keep the focus on the most salient dynamic and temporal aspects of this problem. For example, using the general approach outlined in \citet{zheng2019learning}, it is possible to model complicated nonlinear dependencies via neural networks or orthogonal basis expansions. Furthermore, it is straightforward to replace the least squares loss with the logistic loss (or more generally, any exponential family log-likelihood) to model binary data.
It is also possible to go a step further and consider combinations of continuous and discrete data \citep{andrews2019}, which is important for many real-world applications.

\clearpage
\bibliography{ref}

\clearpage
\appendix
\begin{appendices}
\section*{Appendices}
\section{Comparison of one-stage and two-stage algorithms}\label{sec:oneStageVsTwoStage}

It is possible to minimize the DYNOTEARS objective using either a one-stage algorithm (see \cref{subsec:optimizationProblem}) or a two-stage algorithm (see \cref{subsec:alternativeFormulations}). The two formulations give nearly identical results when the number of samples exceeds the number of variables (i.e., when $n \gg dp$). However, the two-stage algorithm runs somewhat faster, so it should be the preferred option in cases where there is sufficient data.

The difference between the two implementations becomes noticeable especially when the number of samples is below the number of variables.
In such cases, estimating the reduced-form VAR from \cref{eqn:transformedVAR} leads to overfitting. In particular, when $n<dp$, we are solving an underdetermined system, so the residual is $\emat=0$ and we cannot get a meaningful estimate of $\Wmat$. One might resort to regularization by imposing an $\ell_1$ penalty to enforce the sparsity of $\Bmat_i$ in \Cref{eqn:transformedVAR}. As $\Amat_i=\Bmat_i(\Imat-\Wmat)=\Bmat_i-\Bmat_i\Wmat$, we see that the sparsity of $\Bmat_i$ does not translate directly to the sparsity of $\Amat_i$. We also observe empirically that $\Amat_i$ is denser than $\Bmat_i$ in most cases. As a result, one should use a larger regularization parameter in the two-stage setting compared to the one-stage setting. However, to prevent error propagation, it is preferable to estimate $\Wmat$ and $\Amat$ simultaneously via the combined loss from \labelcref{eqn:l1optimizationProblem}.

\section{Numerical experiments}\label{subsec:numericalExperiments}

\subsection{Alternative algorithms}\label{subsec:otherAlgorithms}

The first algorithm that we use for benchmarking is based on an approach that Murphy proposed in \citet{murphy2002thesis}. His idea was to learn intra-slice and inter-slice structures independently; the former task reduces to a static structure-learning problem, and the latter can be viewed as a feature-selection problem.
We use static NOTEARS (with $\ell_1$ regularization) to estimate $\Wmat$, and we use Lasso regression (which incorporates $\ell_1$ regularization by definition) to estimate $\Amat$. This provides a more appropriate comparison to DYNOTEARS than the original setup \citep{friedman1998,murphy2002thesis}. Note that learning $\Wmat$ and $\Amat$ independently in this way is not equivalent to the two-step formulation of DYNOTEARS from \cref{subsec:alternativeFormulations}; in the latter case, we apply NOTEARS to the residuals $\emat$, rather than to the original data $\Xmat$. Our variant of Murphy's method has the same hyperparameters as DYNOTEARS: two regularization parameters $\lambda_{\Wmat}$ and $\lambda_{\Amat}$, and two thresholds $\tau_{\Wmat}$ and $\tau_{\Amat}$ for the weights.

The second algorithm is the SVAR estimation method from \citet{hyvarinen2010}, a time-series version of the LiNGAM algorithm from \citet{shimizu2006}. It follows the two-step approach from \cref{subsec:alternativeFormulations} of first estimating a reduced-form VAR model and then applying LiNGAM to the residuals. With the assumption of non-Gaussian errors, the resulting model is identifiable \citep{hyvarinen2010, shimizu2006}. The two hyperparameters of the time-series version of LiNGAM are the weight thresholds $\tau_{\Wmat}$ and $\tau_{\Amat}$, which we include for comparability with the other algorithms. (The authors of \citealp{hyvarinen2010} did not have this thresholding as part of their method.)

The third algorithm in our experiments is tsGFCI \citep{malinsky2018}, a time-series extension of the Greedy Fast Causal Inference (GFCI) algorithm \citep{ogarrio2016}. Both GFCI and tsGFCI are hybrid algorithms that rely on conditional-independence tests and on local changes to a graph to incrementally improve the BIC score. These algorithms work in settings with latent variables and return a partial ancestral graph (PAG).
We define heuristics to extract adjacency matrices $\Wmat$ and $\Amat$ from the output PAG (see \cref{subsec:PagToDag} for details). When there is ambiguity in whether an edge is present or not, we treat tsGFCI as favorably as possible. One important parameter in tsGFCI is the ``penalty discount"; larger values of this parameter increase the BIC penalty and thus result in sparser output graphs. In our experiments on simulated data, we find that setting the penalty discount between $2$ and $4$ produces output graphs that are closest to the ground truth. Our simulations from \cref{subsec:simulationResults} and \cref{subsec:additionalResults} use a value of $2$ (which is also the default value). %

\subsection{Interpreting a PAG as a DAG}\label{subsec:PagToDag}

The tsGFCI algorithm returns a partial ancestral graph (PAG), which one cannot immediately compare to a ground-truth DAG. Thus, we developed a set of rules to convert the PAG output to a DAG, making sure to do so in a manner that favors tsGFCI. The rules are as follows:
\begin{enumerate}[itemsep=-2pt,before=\vspace{-1em},after=\vspace{-1em},label={$\bullet$}]
\item If an edge is directed (i.e., A $\rightarrow$ B in the PAG), then we treat it as a directed edge in the DAG.
\item If an edge in the PAG is either directed or it indicates the presence of a latent factor (i.e., A $\circ\rightarrow$ B), then we check whether the directed edge exists in the ground truth graph and assume that tsGFCI made the correct choice.
\item If two nodes are related through a latent variable (i.e., A $\longleftrightarrow$ B in the PAG), then we disregard the edge.
\item If the edge is ambiguous (i.e., A $\circ - \circ$ B), then we assume that tsGFCI made the correct choice; in other words, we check whether A $\rightarrow$ B, B $\leftarrow$ A, or A is not connected to B in the ground-truth DAG and we assume that tsGFCI made the same choice.
\end{enumerate}
Using these rules, we pick the outcomes most favorable for tsGFCI in ambiguous cases. This implies, in particular, that our results slightly overstate the performance of tsGFCI on simulated data.

\subsection{Hyperparameter selection}\label{subsec:hyperparameters}

For our simulations with $n=500$, we apply a small amount of regularization, $\lambda_{\Wmat}=\lambda_{\Amat}=0.05$, for both DYNOTEARS and NOTEARS + Lasso. Because the number of samples exceeds the number of variables, performance is not particularly sensitive to the amount of regularization. For all algorithms except tsGFCI, we apply the weight thresholds $\tau_{\Wmat}=0.3$ and $\tau_{\Amat}=0.1$. We set $\tau_{\Wmat}=0.3$ to be consistent with the experiments for static NOTEARS from \citet{zheng2018}, and we set $\tau_{\Amat}=0.1$ using an analogous heuristic. (We experimented with other threshold values, and the relative ranking of the algorithms was largely the same.) For tsGFCI, we set the penalty discount to $2$; recall discussion from \cref{subsec:otherAlgorithms}.

Regularization becomes more important for $n=50$, so we set $\lambda_{\Wmat}=\lambda_{\Amat}=0.2$ for DYNOTEARS and NOTEARS + Lasso. We set $\tau_{\Wmat}=0.3$ and $\tau_{\Amat}=0.2$. We keep the penalty discount at $2$ for tsGFCI, as experimentation with other values did not yield superior results.

Although we did not attempt to optimize hyperparameters for our experiments on simulated data, our work on the S\&P100 and the DREAM4 datasets (see \Cref{sec:applications}) indicates ways in which one can estimate these parameters through cross-validation.

\subsection{Data generation process}\label{subsec:dataGenProcess}

We provide more details about the data generation process that we use in our numerical experiments from \Cref{sec:numericalExperiments}.

\paragraph{Intra-slice model}
As in \citet{zheng2018}, we use either the Erd\H{o}s--R\'{e}nyi (ER, \citealp{newman2018}) model or the Barab\'{a}si--Albert (BA, \citealp{barabasi1999}) model to generate intra-slice graphs given a target mean degree $k$ (which counts both incoming and outgoing edges). In the ER model, one samples edges using i.i.d. Bernoulli trials. To ensure that the resulting graph is a DAG, we sample lower-triangular entries of $\Wmat^\mathrm{bin}$ in this way, and we then permute the rows and columns to randomize node order. By setting the probability of each Bernoulli trial to $k / (d-1)$, the expected mean degree in the resulting graph is $k$, as desired. The BA model \citep{barabasi1999} relies on a ``preferential attachment" mechanism to generate growing networks in which nodes are added one by one. For each new node, one generates $k / 2$ outgoing edges. The targets of these edges are selected at random from the existing nodes, proportionally to their current degrees. This mechanism encapsulates a ``rich get richer" effect that produces, at the end of the process, graphs with a power-law degree distribution. The BA model thus produces graphs that mirror the wide degree distributions that are common in many real-world networks. By construction, this formulation of the BA model generates DAGs. As for ER models, we permute the rows and columns of the resulting adjacency matrix so that nodes are not a priori sorted in the topological order.

To go from an unweighted to a weighted DAG, we follow \citet{zheng2018} and we sample weights uniformly at random from $[-2.0,\,-0.5] \cup [0.5,\,2.0]$.

\paragraph{Inter-slice model}
We use two models to generate inter-slice graphs. One is a directed ER model in which we sample entries of the binary adjacency matrix $\Amat^{\mathrm{bin}}$ using i.i.d. Bernoulli trials with probabilities $k / d$. This choice implies that the mean in-degree of nodes at time $t$ is equal to $pk$. The other model is a simplified type of the so-called stochastic block model (SBM,  \citealp{newman2018}). In our simulations, we partition the $d$ variables into two blocks, and we assume that the probability of an inter-slice edge from $i$ to $j$ is equal to $p_\mathrm{in}$ if $i$ and $j$ are in the same block, and it is equal to $p_\mathrm{out}$ otherwise. We select $p_\mathrm{in}$ and $p_\mathrm{out}$ such that $p_\mathrm{out}/p_\mathrm{in}=0.3$ (so edges are more likely between two variables in the same cluster) and such that the expected mean in-degree of nodes at time $t$ is $pk$. In applications to real data we expect some clustering of the variables and of their causal effects, and we rely on SBMs to replicate this feature in our simulation experiments.

Given a binary inter-slice adjacency matrix $\Amat^{\mathrm{bin}}$, we sample edge weights uniformly at random from a specified interval, which we allow to depend on $p$. More precisely, we sample edge weights from slice $t-p$ to slice $t$ from $[-0.5\alpha,\,-0.3\alpha] \cup [0.3\alpha,\,0.5\alpha]$, where $\alpha=1/\eta^{p-1}$ with $\eta \geq 1$. The weight decay parameter $\eta$ therefore reduces the influence of variables that are farther back in time from the current time slice. (Of course, there is no such penalty when $\eta=1$.) We incorporate this parameter into our data-generation process because it replicates a time-decay scenario that we expect to encounter in real-world applications. It also allows us to determine whether DYNOTEARS can learn edge weights across multiple scales, ideally without having to specify different thresholds for each matrix $\Amat_i$ ($i \in \{1,\ldots,p\}$). %

Once we have $\Wmat$ and $\Amat$, we use the SEM from \labelcref{eqn:dynSEMcompact} to generate a data matrix $\Xmat$ of size $n \times d$. %
The noise term $\Zmat$ in \labelcref{eqn:dynSEMcompact} is a matrix of i.i.d.\ random variables. Following \citet{zheng2018}, we use normal and exponential distributions with tuneable scale parameters (set to $1$ by default) for these random variables.

\begin{figure}[h!]
	\centering
	\begin{subfigure}{0.8\columnwidth}
		\includegraphics[width=\textwidth]{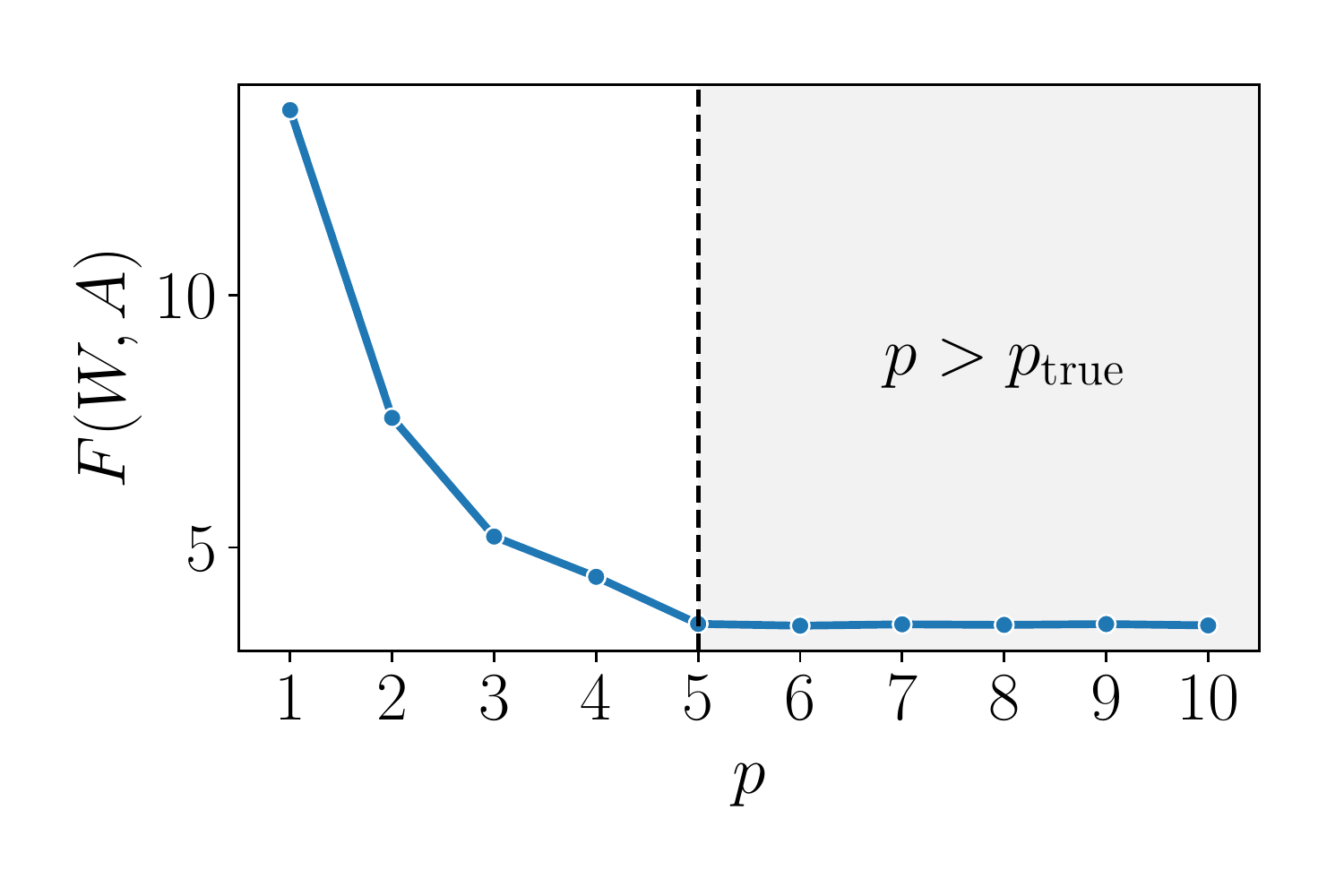}
		\caption{Objective value as a function of $p$}
		\label{fig:vary_p_F_vals}
	\end{subfigure}

	\begin{subfigure}{0.8\columnwidth}
		\includegraphics[width=\textwidth]{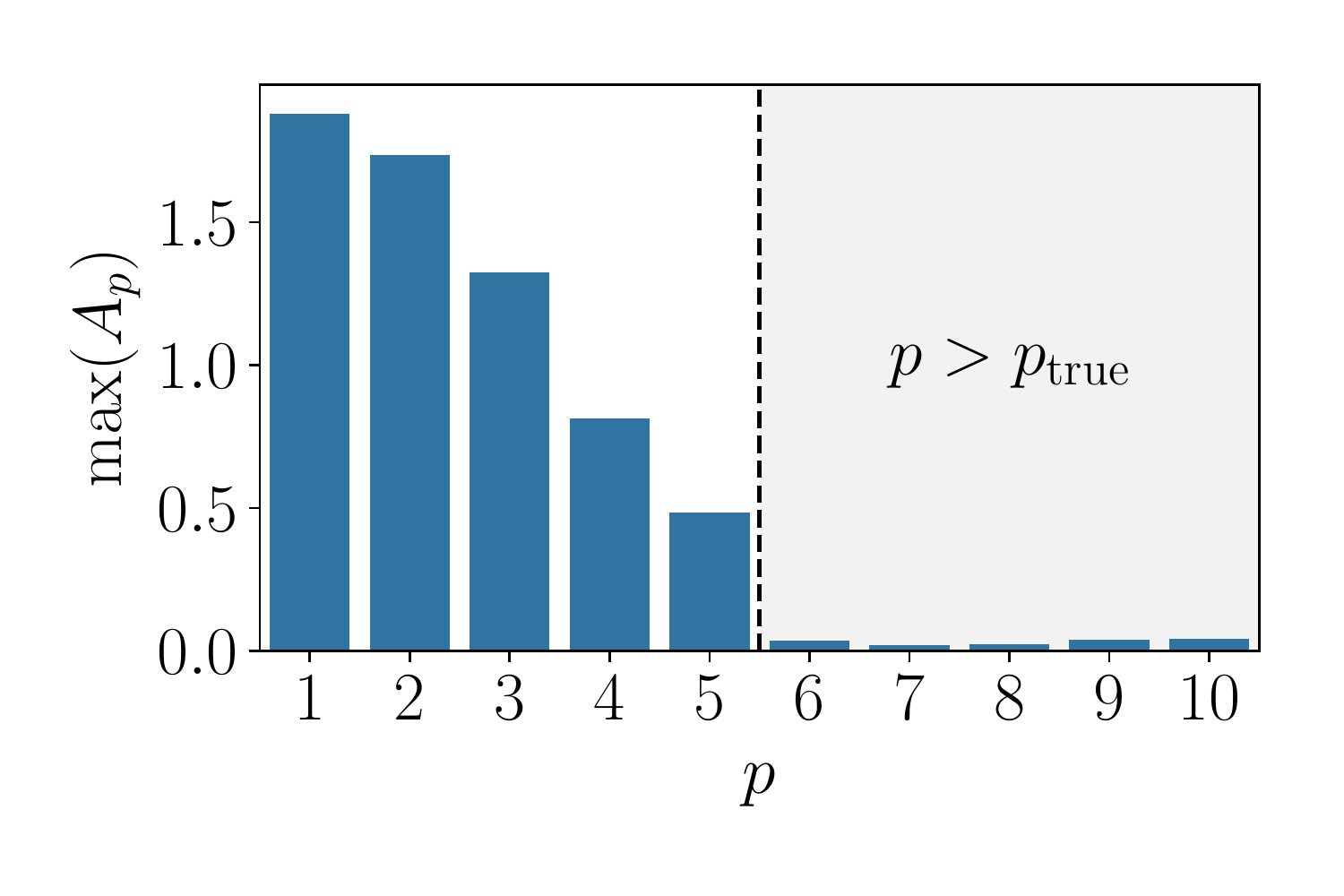}
		\caption{Largest absolute value in $\Amat_p$ as a function of $p$}
		\label{fig:vary_p_max_Ap}
	\end{subfigure}
	\caption{Results for fitting a model using different values of $p$ to data with $p_\mathrm{true}=5$. (a) The objective value $F(\Wmat,\Amat)$ plateaus for $p > p_\mathrm{true}$. (b) The estimated edge weights in $\Amat_p$ are close to $0$ for $p > p_\mathrm{true}$.}
	\label{fig:vary_p}
\end{figure}

\subsection{Autoregressive order}\label{subsec:autoregressiveOrder}

In \cref{subsec:simulationResults}, we assumed that the correct value of the autoregressive order $p$ is given. This is rarely the case in applications, so it is useful to be able to estimate $p$ from data. In \cref{fig:vary_p}, we indicate two such potential diagnostics for a simulated dataset with $p_\mathrm{true}=5$. In \cref{fig:vary_p_F_vals}, we see that the objective function decreases as we run DYNOTEARS for increasing values of $p$. However, the objective values plateau once $p>p_\mathrm{true}$, as the increasingly complex models do not yield better fits to the data. For real-world data, where there is no single ``true" value for $p$, one can also look at plateaus in the BIC score. An alternative method for selecting $p$ is to look at the magnitude of the weights in the estimated inter-slice matrices, as we do in \cref{fig:vary_p_max_Ap}. For $p > p_\mathrm{true}$, $\Amat_p$ does not contain entries that are significantly above $0$ in magnitude. Thus, in cases when $p$ is unknown, one can keep increasing $p$ until the entries of $\Amat_p$ become negligible.

\subsection{Running times}

\begin{figure}[ht!]
	\centering
	\includegraphics[width=\columnwidth]{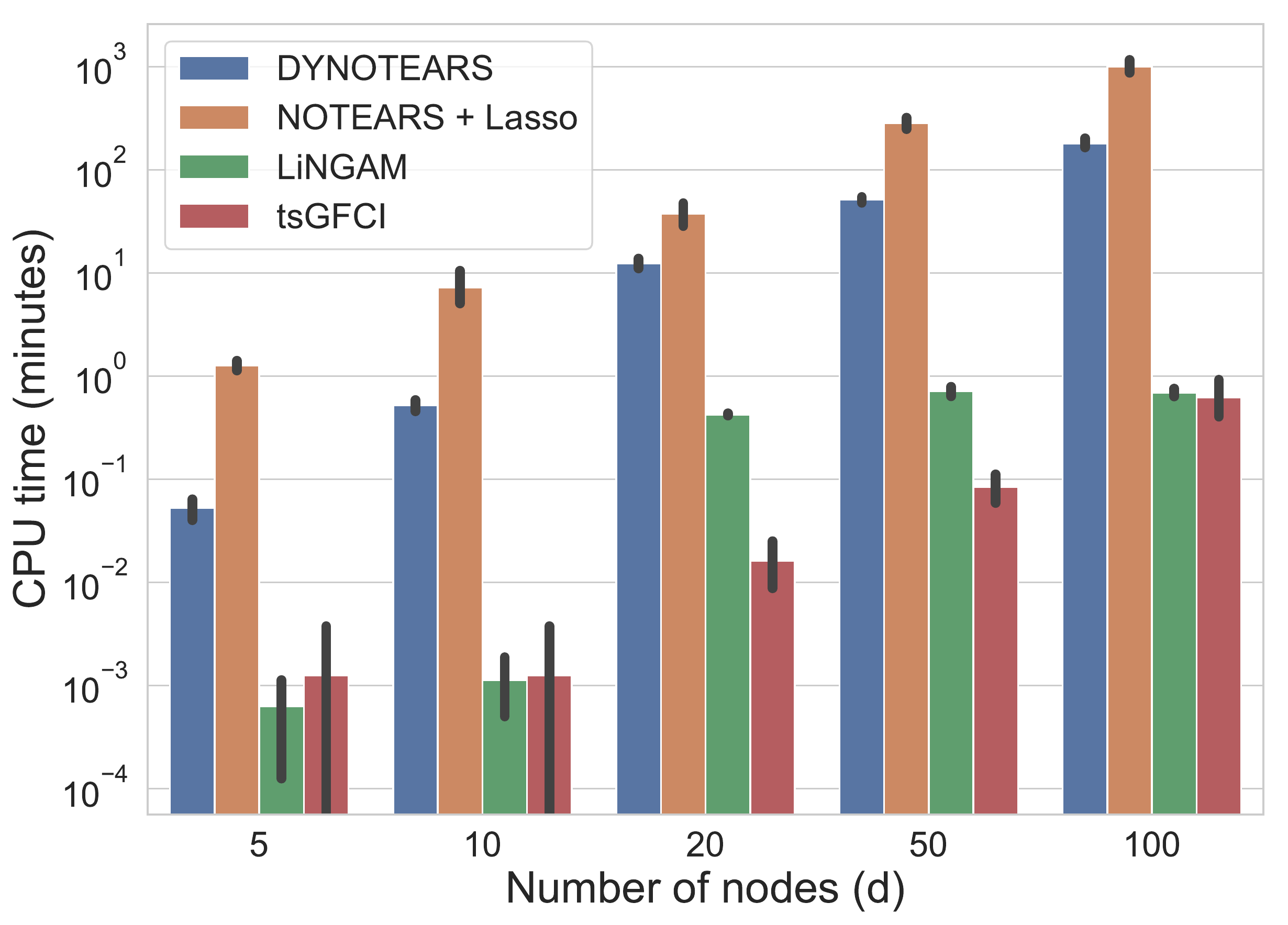}
	\caption{CPU times for the simulations from \cref{fig:simGaussF1} (Gaussian noise and $n=500$). Running times are comparable for Exponential noise and for $n=50$.}
	\label{fig:runningTimes}
\end{figure}

We provide some illustrative running times for different values of $d$ in \cref{fig:runningTimes}. These running times also depend on the density of the underlying graph and on the distribution of edge weights, although it is difficult to quantify the precise relationship. The speed of DYNOTEARS and NOTEARS + LiNGAM is heavily dependent on the values of the regularisation parameters, with larger values resulting in faster running times.\footnote{For the DREAM4 datasets, the CPU times are approximately $0.1$ minutes, $1.5$ minutes, and $60$ minutes, respectively, for $\lambda_{\Wmat} \in \{0.1, 0.01, 0.001\}$ and $\lambda_{\Amat}=0.01$.}

Although tsGFCI and LiNGAM run significantly faster than DYNOTEARS, we believe that the gain in accuracy from using the latter makes it worthwhile even in cases with hundreds of variables. As a reminder, running DYNOTEARS on the S\&P100 dataset (which has $d \approx 100$) takes a few minutes on a typical laptop.

\subsection{Additional results}\label{subsec:additionalResults}

\begin{figure*}[p!]
	\centering
	\includegraphics[width=\textwidth]{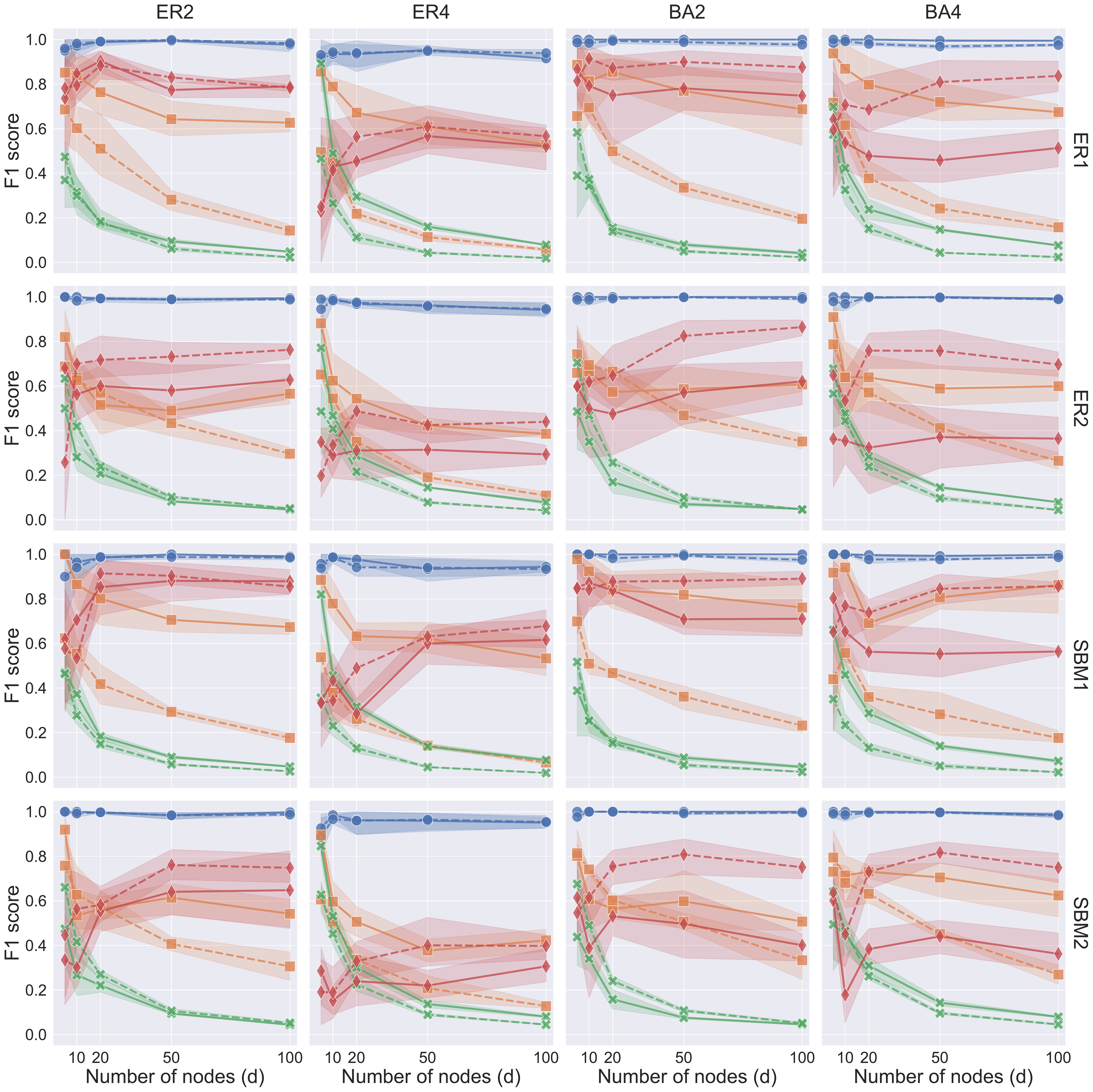} \\
	\includegraphics[width=\textwidth]{sim_legend_4alg.pdf}
	\caption{F1 score for $n=500$, $p=1$, $d \in \{5, 10, 20, 50, 100\}$, Gaussian noise, and different choices of intra-slice graphs (columns) and inter-slice graphs (rows). Each marker indicates the mean performance across $5$ algorithms runs (each on a different simulated dataset).}
	\label{fig:simGaussF1}
\end{figure*}

\paragraph{Additional ground-truth graphs}
The relative performance of different algorithms varies as we change the density of the ground-truth graphs. In \cref{fig:simGaussF1}, we show the F1 scores for simulated data with Gaussian noise, $n=500$, four choices of intra-slice graphs (columns), and four choices of inter-slice graphs (rows). The performance of tsGFCI is especially sensitive to changes in graph densities, with a notable drop in F1 scores when the intra-slice graph is ER4.

\begin{figure*}[p!]
	\centering
	\begin{subfigure}{\columnwidth}
		\includegraphics[width=\textwidth]{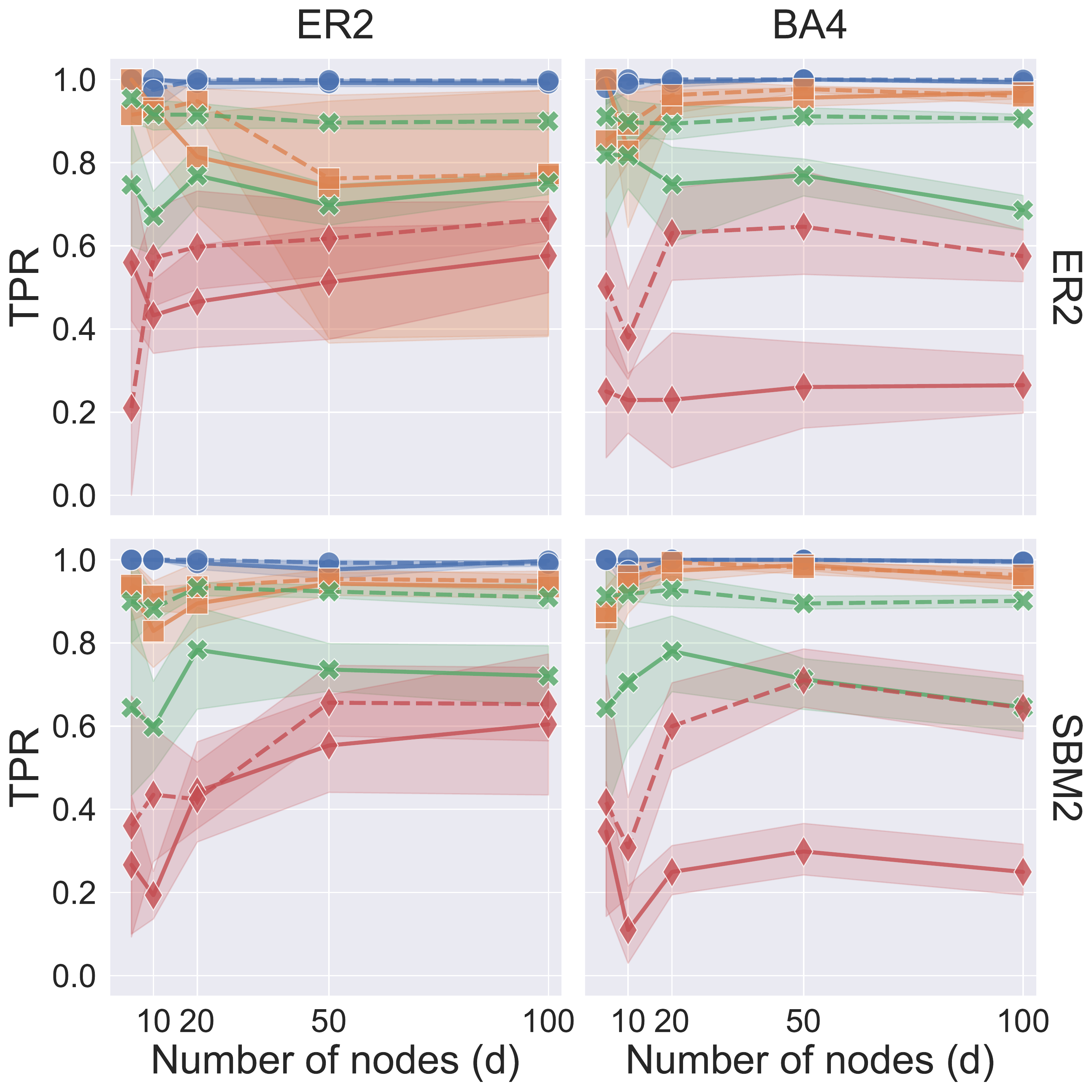}
		\caption{True positive rate}
		\label{fig:simGaussN50tpr}
	\end{subfigure} \hfill
	\begin{subfigure}{\columnwidth}
		\includegraphics[width=\textwidth]{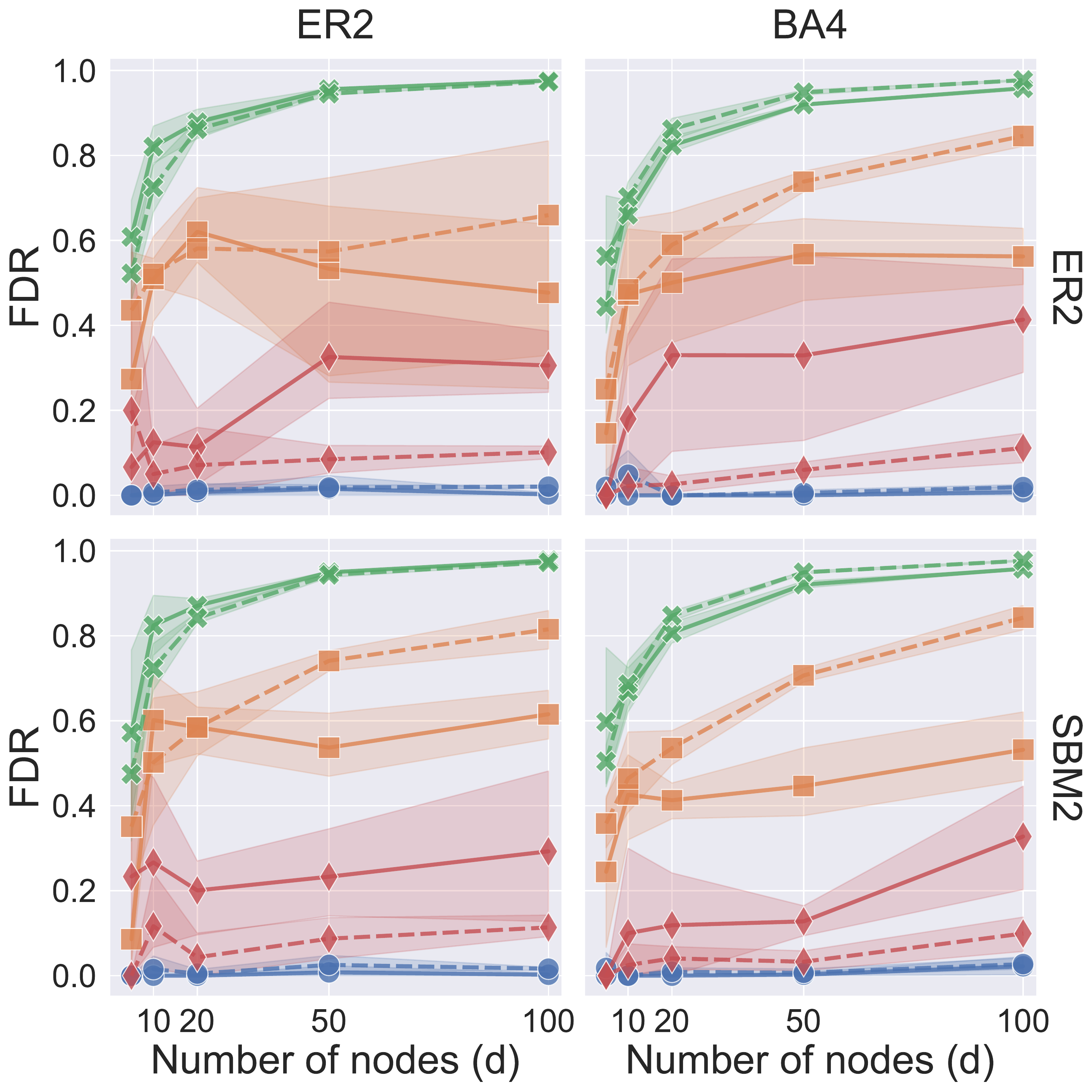}
		\caption{False discovery rate}
		\label{fig:simGaussN50fdr}
	\end{subfigure} \\ \vspace{0.5em}
	\begin{subfigure}{\columnwidth}
		\includegraphics[width=\textwidth]{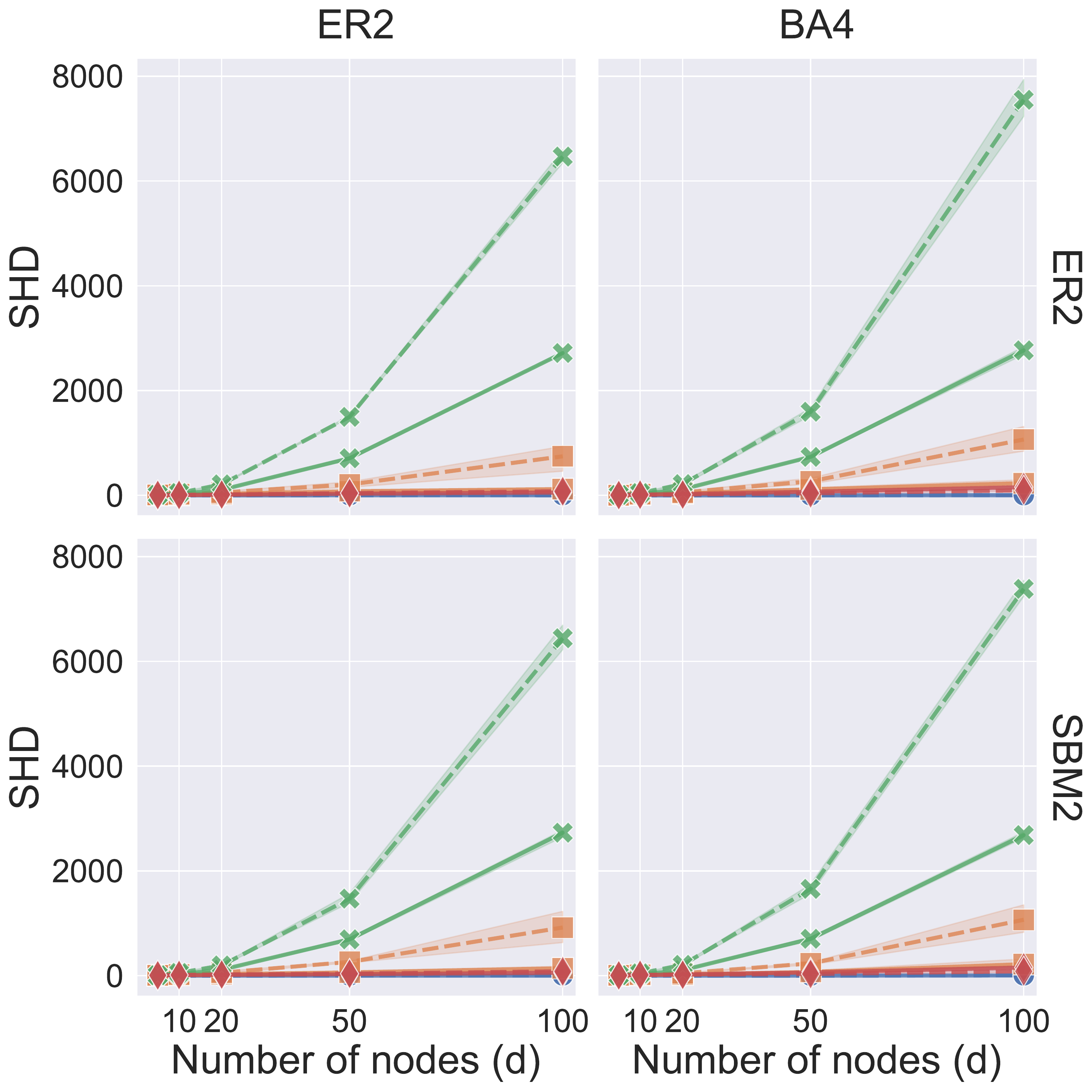}
		\caption{Structural Hamming distance}
		\label{fig:simGaussN50f1}
	\end{subfigure} \hfill
	\begin{subfigure}{\columnwidth}
		\includegraphics[width=\textwidth]{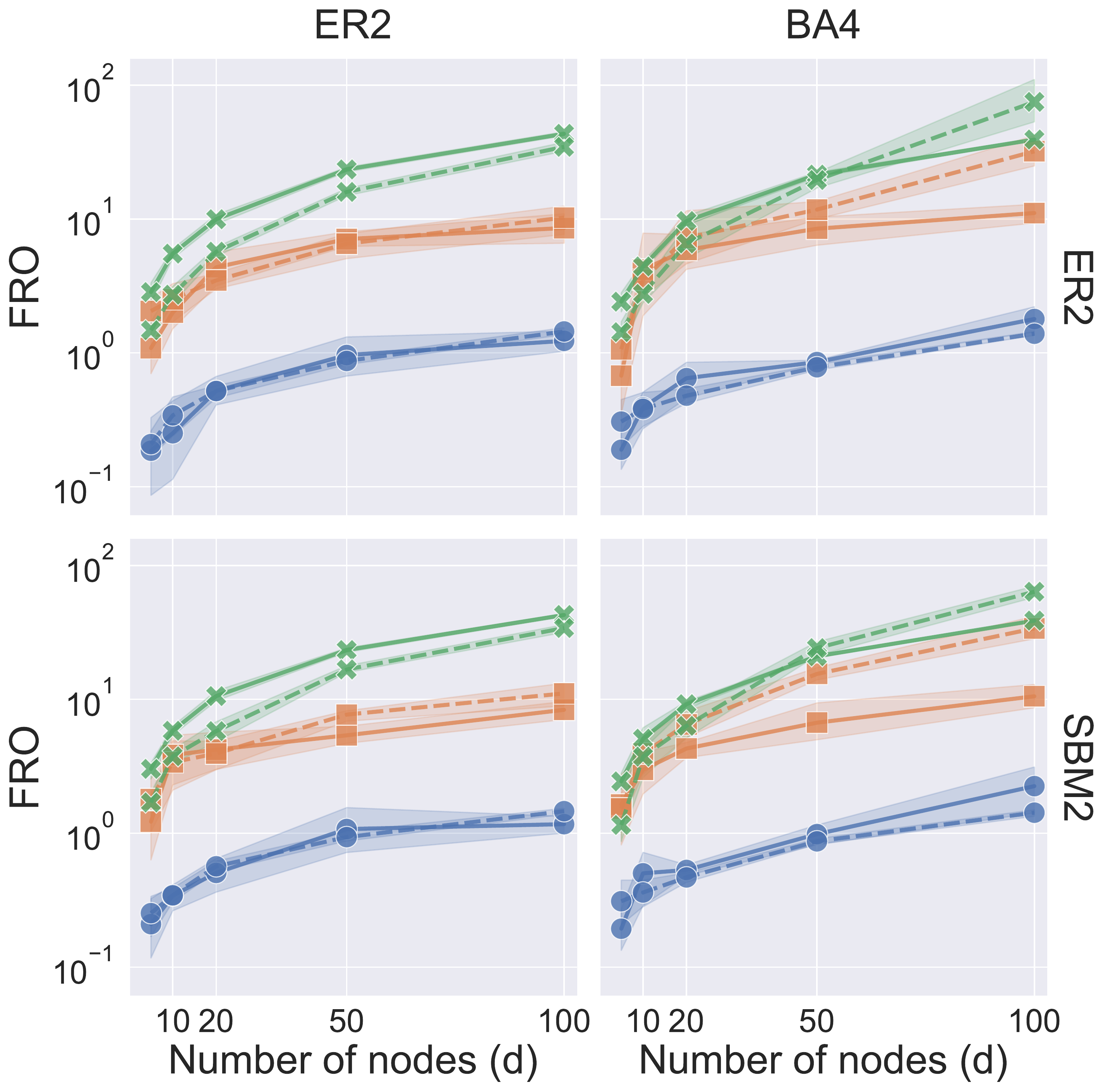}
		\caption{Frobenius norm}
		\label{fig:simGaussN50fro}
	\end{subfigure} \\
	\includegraphics[width=\textwidth]{sim_legend_4alg.pdf}
	\caption{Results for $n=500$, Gaussian noise. Each panel corresponds to a different performance metric.}
	\label{fig:simGaussN500}
\end{figure*}

\begin{figure*}[p!]
	\centering
	\begin{subfigure}{\columnwidth}
		\includegraphics[width=\textwidth]{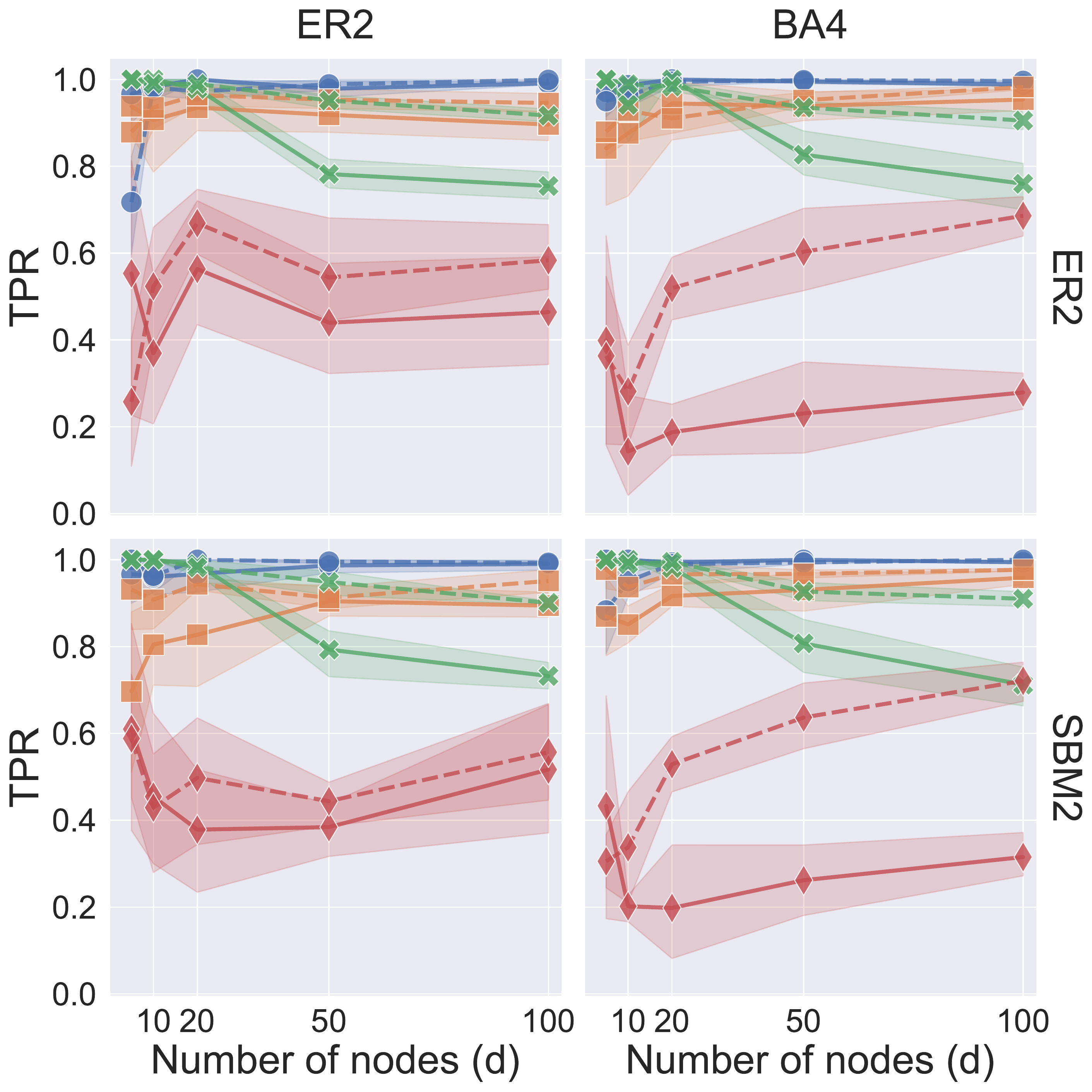}
		\caption{True positive rate}
		\label{fig:simExpN50tpr}
	\end{subfigure} \hfill
	\begin{subfigure}{\columnwidth}
		\includegraphics[width=\textwidth]{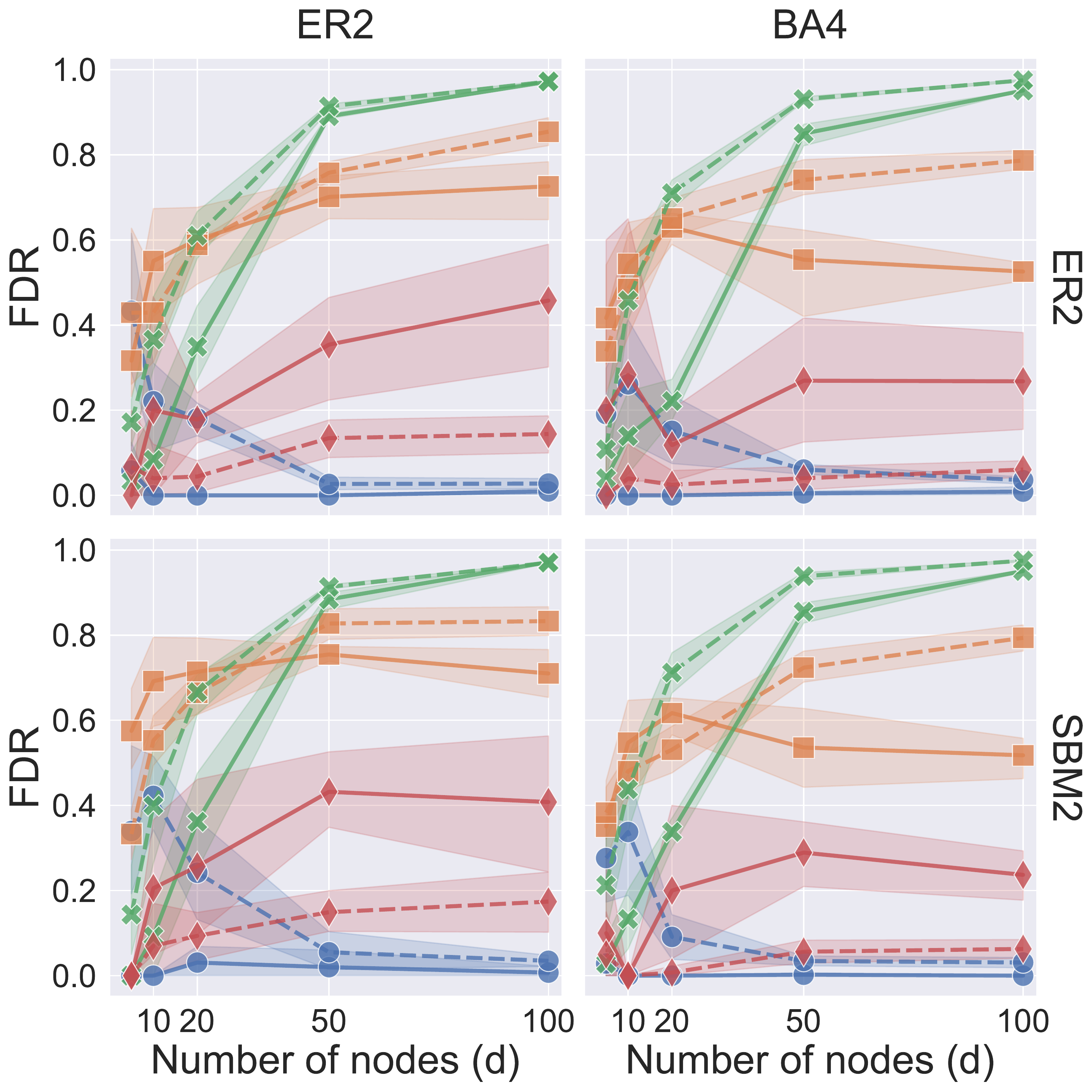}
		\caption{False discovery rate}
		\label{fig:simExpN50fdr}
	\end{subfigure} \\ \vspace{0.5em}
	\begin{subfigure}{\columnwidth}
		\includegraphics[width=\textwidth]{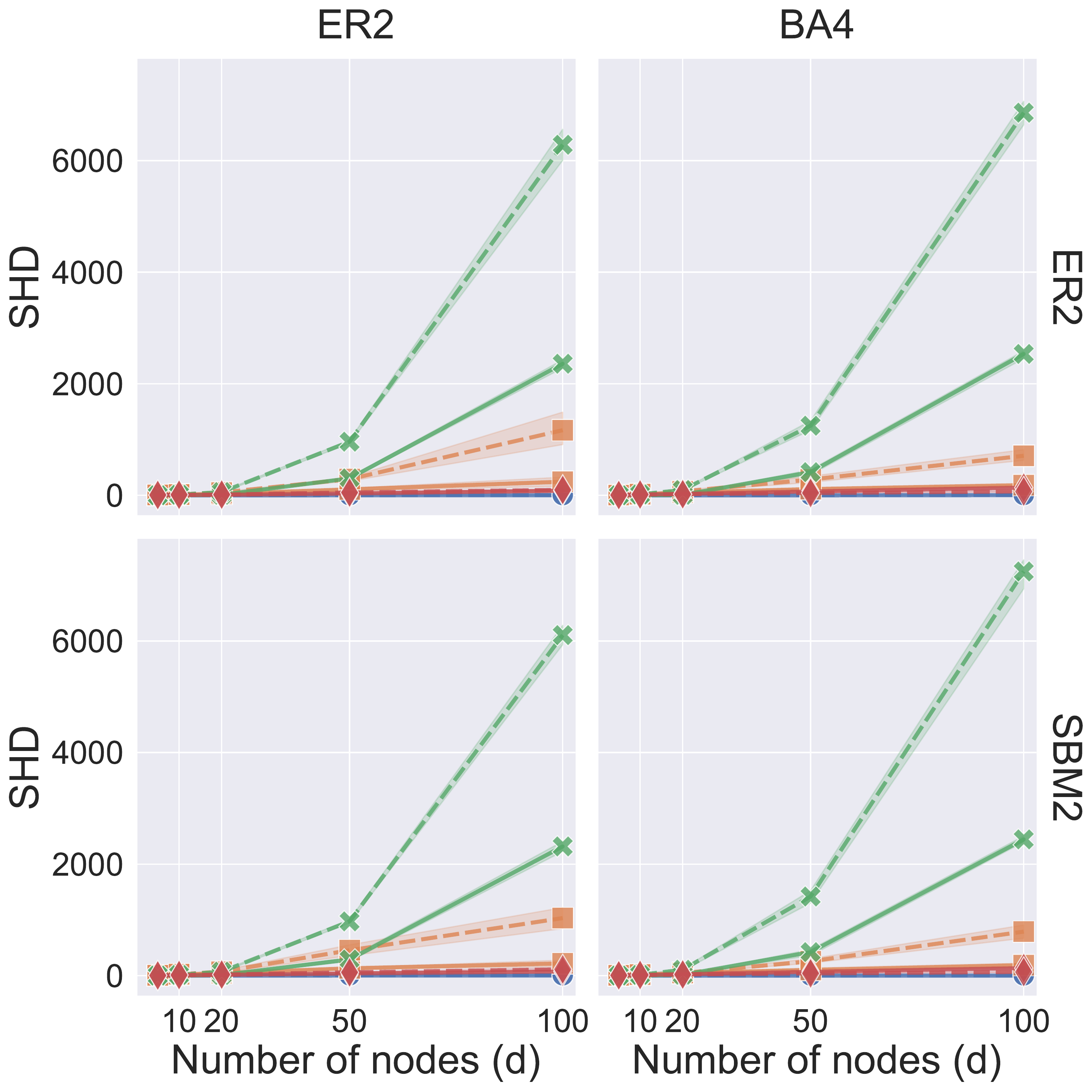}
		\caption{Structural Hamming distance}
		\label{fig:simExpN50f1}
	\end{subfigure} \hfill
	\begin{subfigure}{\columnwidth}
		\includegraphics[width=\textwidth]{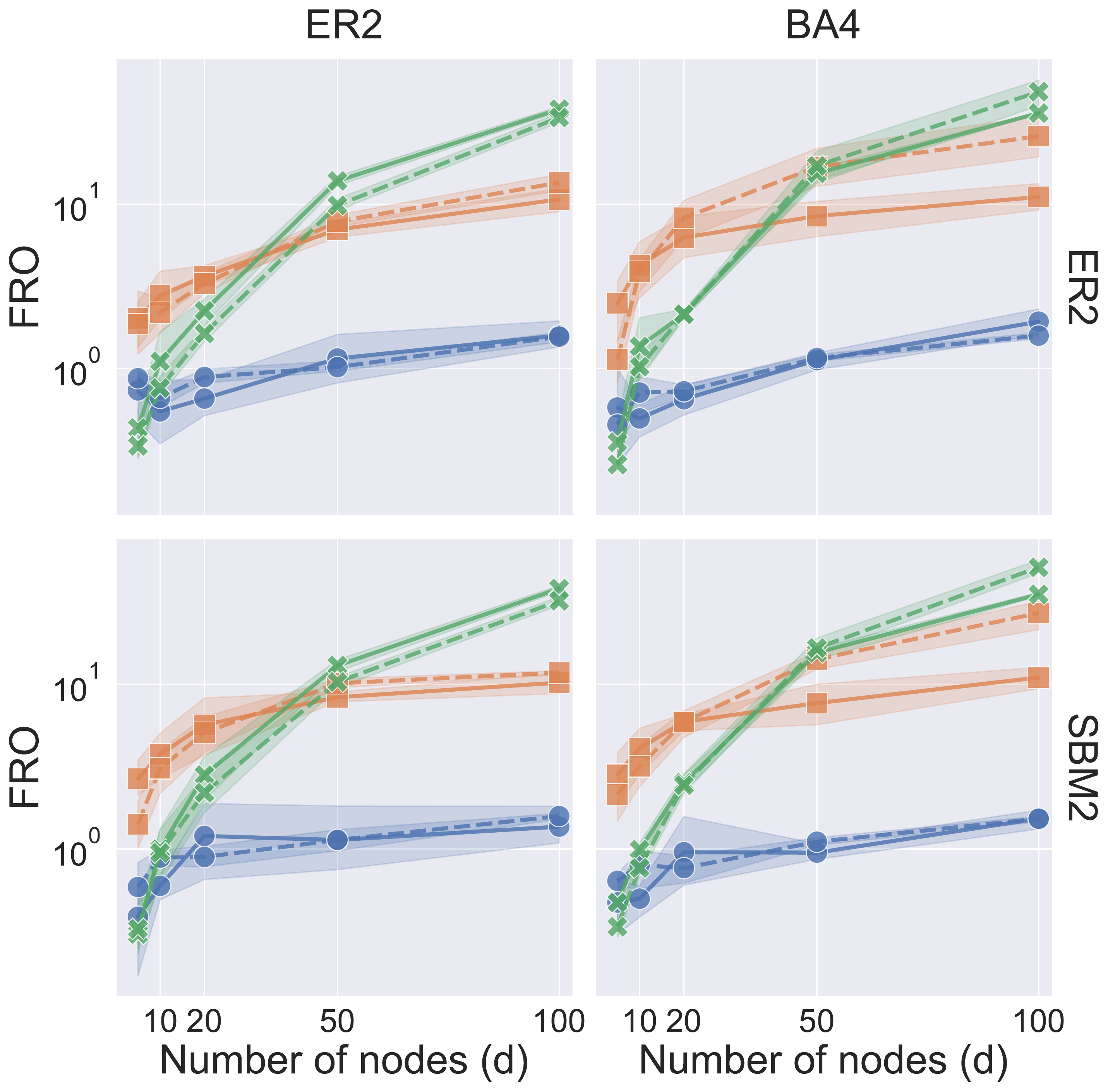}
		\caption{Frobenius norm}
		\label{fig:simExpN50fro}
	\end{subfigure} \\
	\includegraphics[width=\textwidth]{sim_legend_4alg.pdf}
	\caption{Results for $n=500$, exponential noise. Each panel corresponds to a different performance metric.}
	\label{fig:simExpN500}
\end{figure*}

\begin{figure*}[p!]
	\centering
	\begin{subfigure}{\columnwidth}
		\includegraphics[width=\textwidth]{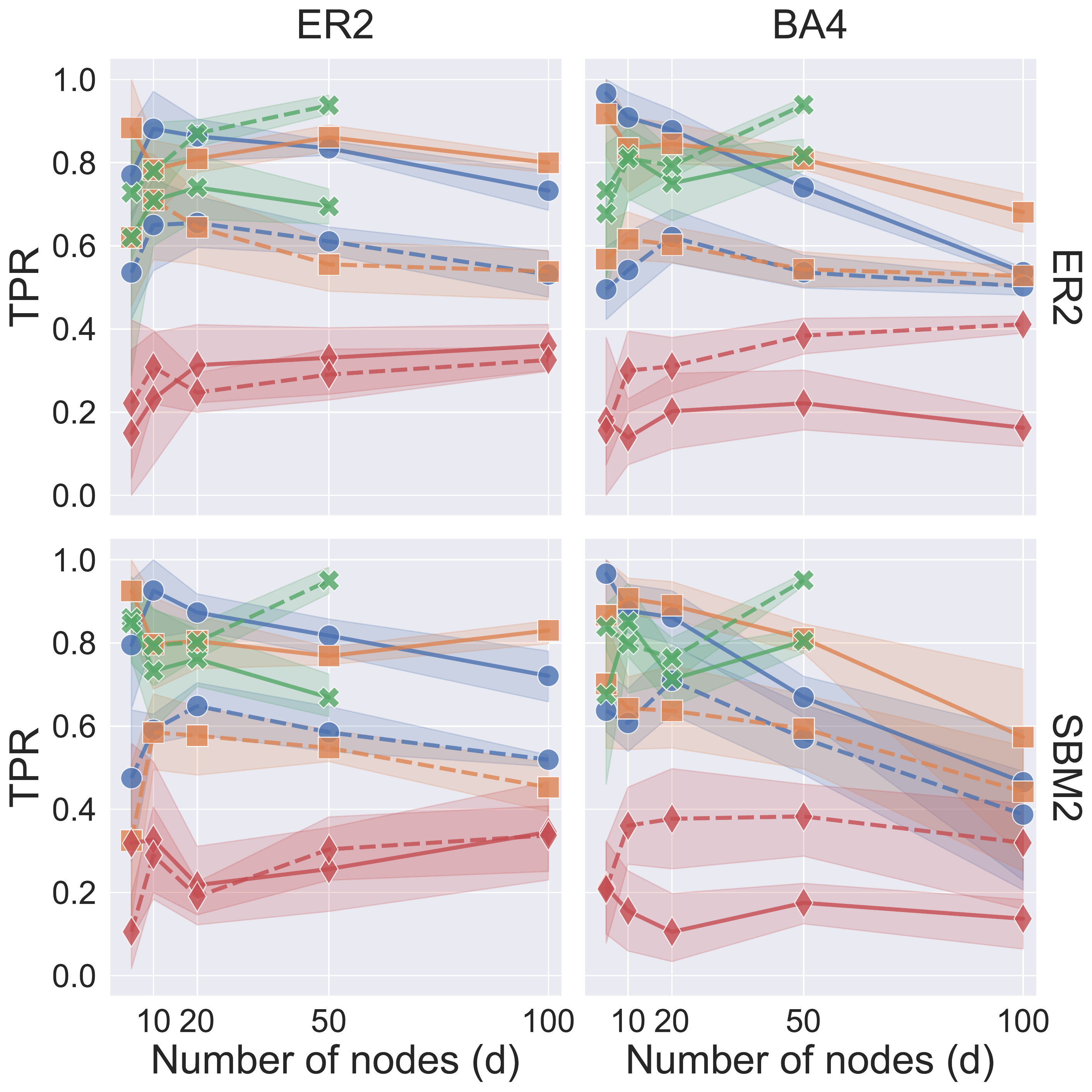}
		\caption{True positive rate}
		\label{fig:simGaussN50tpr}
	\end{subfigure} \hfill
	\begin{subfigure}{\columnwidth}
		\includegraphics[width=\textwidth]{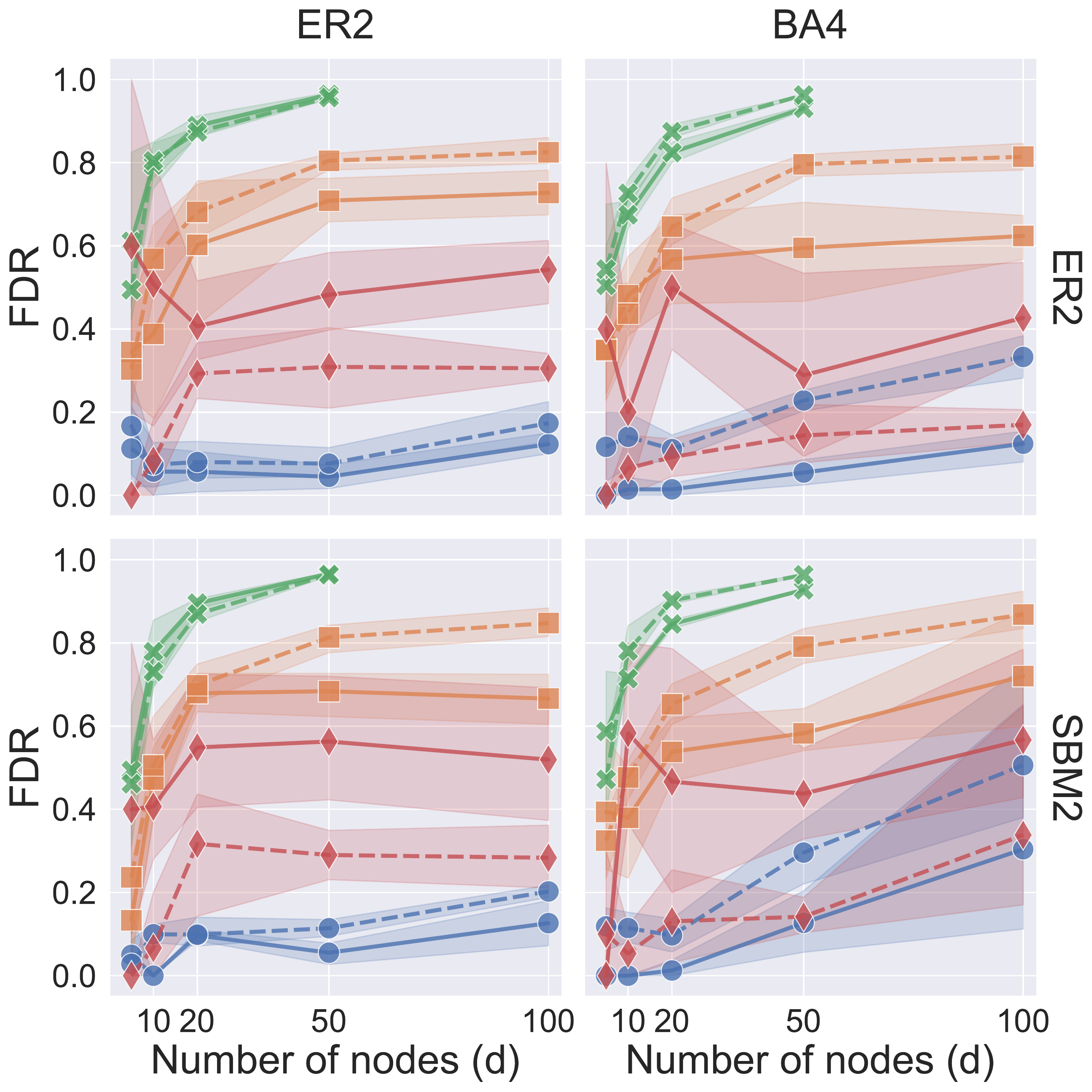}
		\caption{False discovery rate}
		\label{fig:simGaussN50fdr}
	\end{subfigure} \\
	\begin{subfigure}{\columnwidth}
		\includegraphics[width=\textwidth]{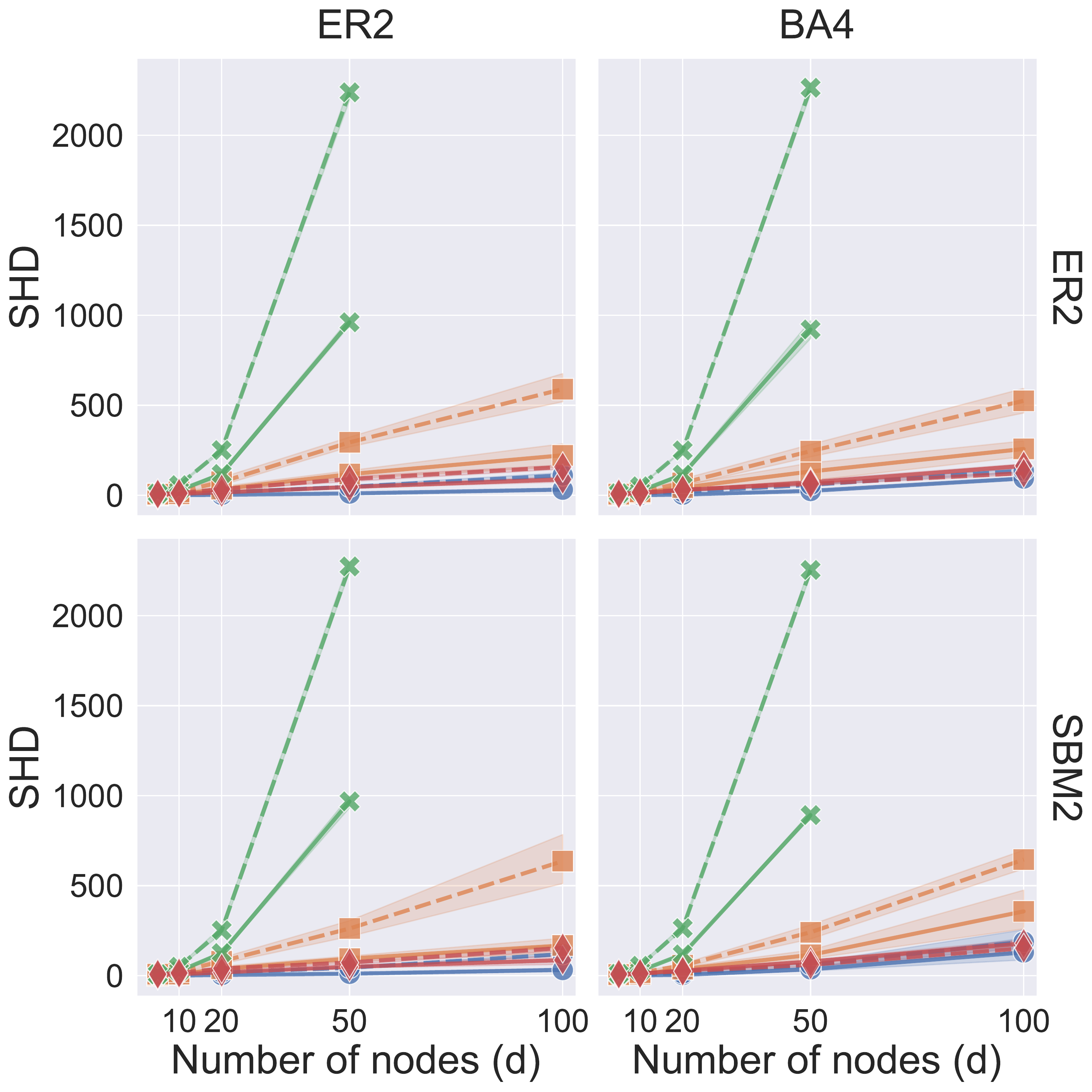}
		\caption{Structural Hamming distance}
		\label{fig:simGaussN50f1}
	\end{subfigure} \hfill
	\begin{subfigure}{\columnwidth}
		\includegraphics[width=\textwidth]{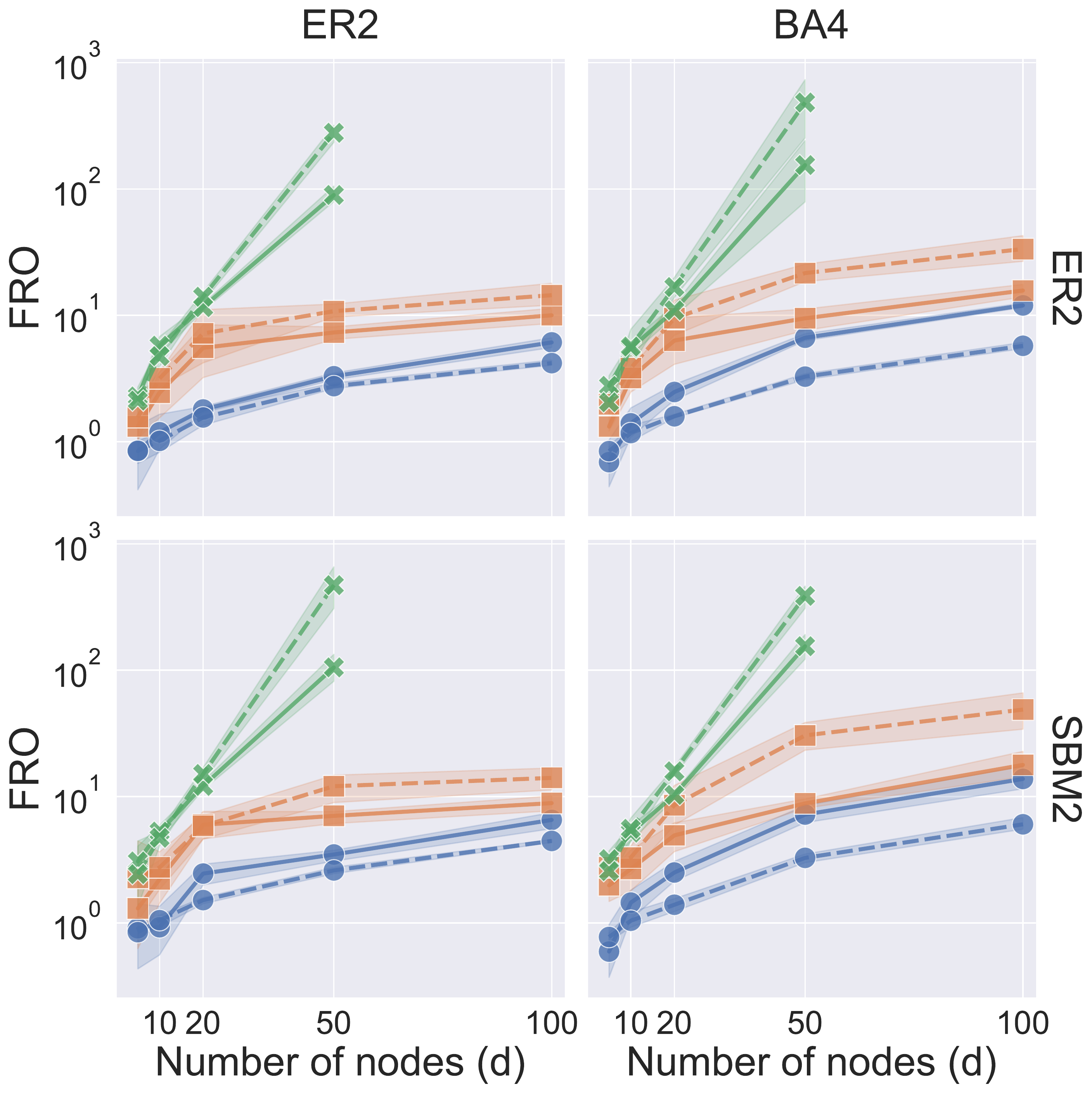}
		\caption{Frobenius norm}
		\label{fig:simGaussN50fro}
	\end{subfigure} \\
	\includegraphics[width=\textwidth]{sim_legend_4alg.pdf}
	\caption{Results for $n=50$, Gaussian noise. Each panel corresponds to a different performance metric. LiNGAM does not work for $n < d$, so the corresponding results are missing from the plots.}
	\label{fig:simGaussN50}
\end{figure*}

\begin{figure*}[p!]
	\centering
	\begin{subfigure}{\columnwidth}
		\includegraphics[width=\textwidth]{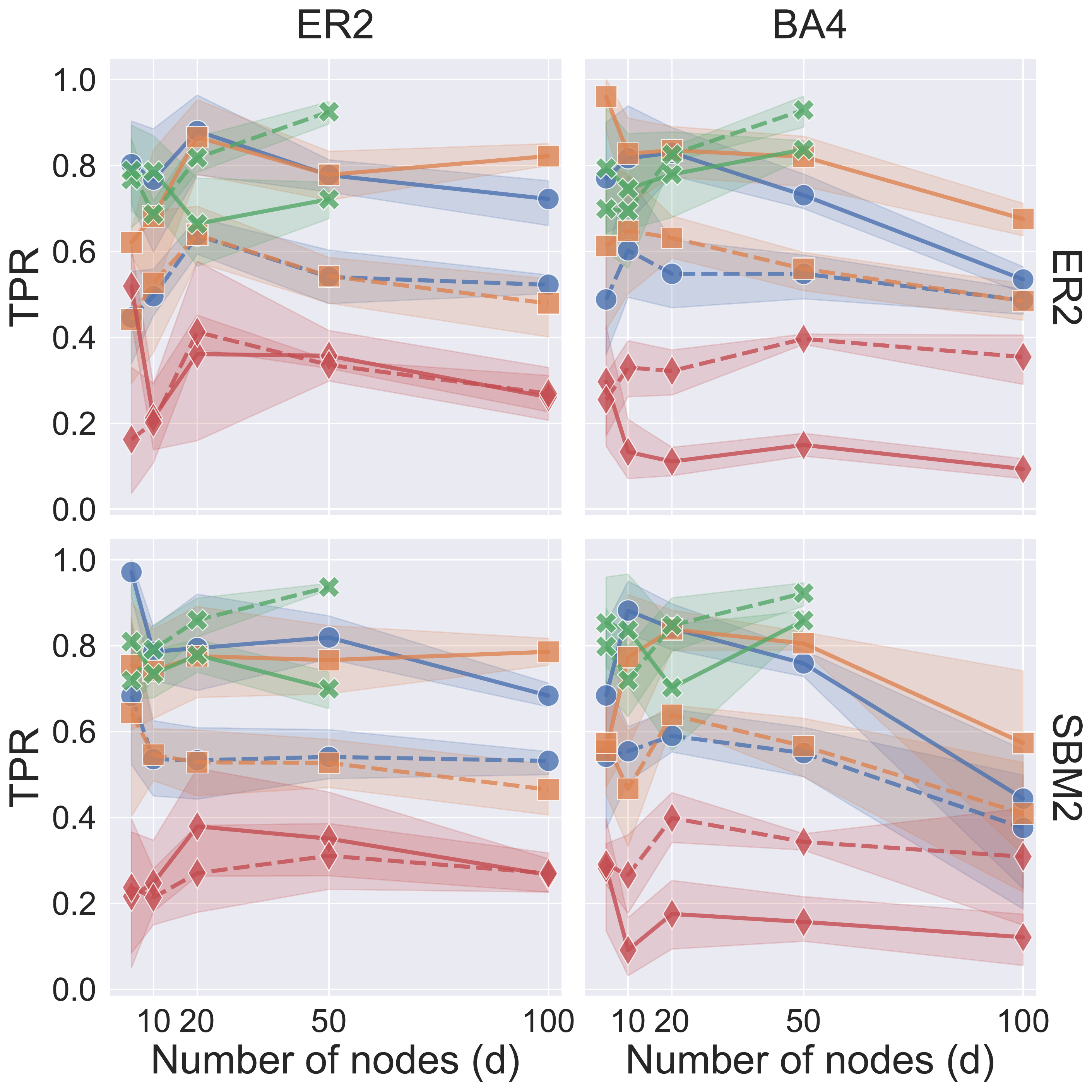}
		\caption{True positive rate}
		\label{fig:simExpN50tpr}
	\end{subfigure} \hfill
	\begin{subfigure}{\columnwidth}
		\includegraphics[width=\textwidth]{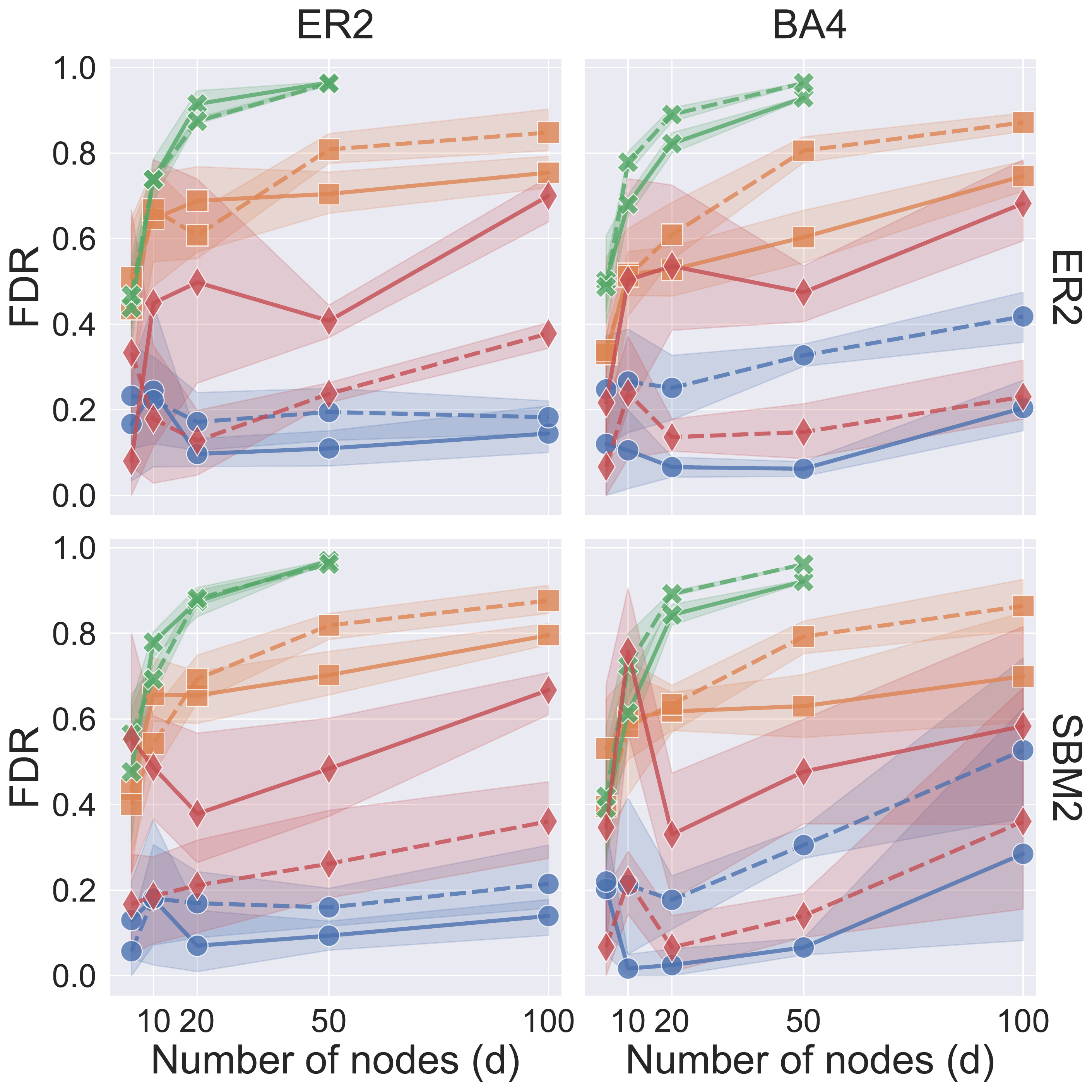}
		\caption{False discovery rate}
		\label{fig:simExpN50fdr}
	\end{subfigure} \\
	\begin{subfigure}{\columnwidth}
		\includegraphics[width=\textwidth]{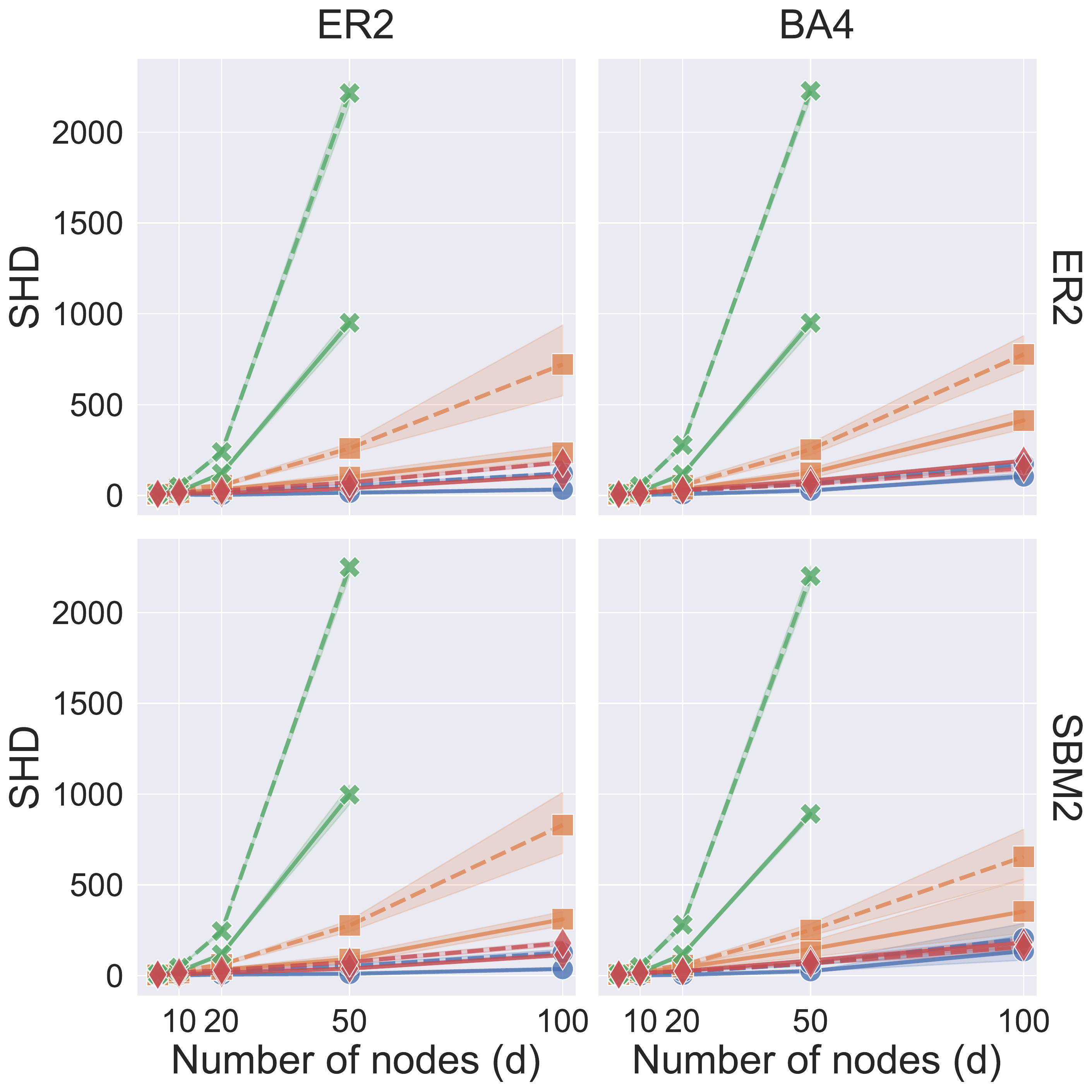}
		\caption{Structural Hamming distance}
		\label{fig:simExpN50f1}
	\end{subfigure} \hfill
	\begin{subfigure}{\columnwidth}
		\includegraphics[width=\textwidth]{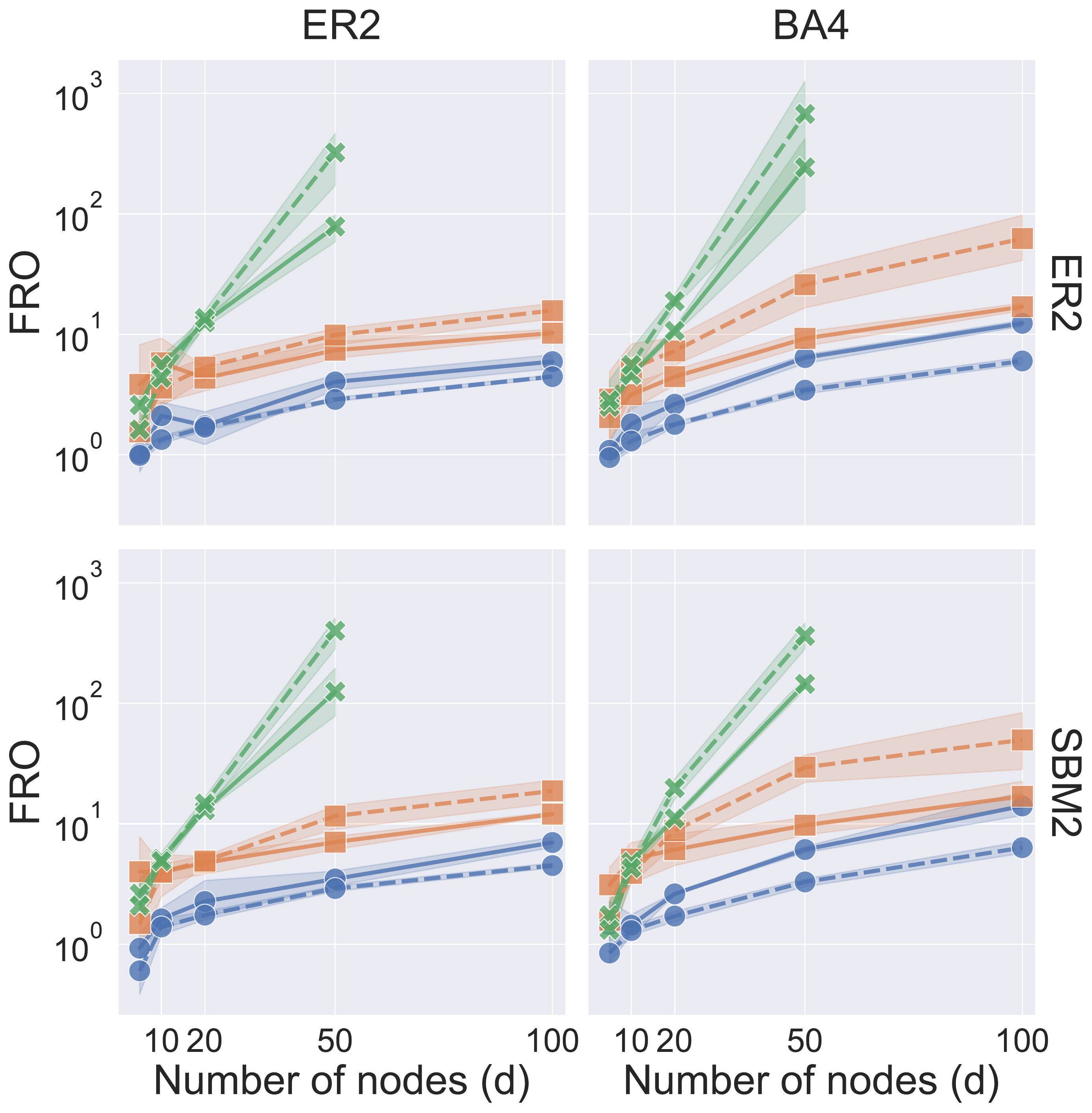}
		\caption{Frobenius norm}
		\label{fig:simExpN50fro}
	\end{subfigure} \\
	\includegraphics[width=\textwidth]{sim_legend_4alg.pdf}
	\caption{Results for $n=50$, exponential noise. Each panel corresponds to a different performance metric. LiNGAM does not work for $n < d$, so the corresponding results are missing from the plots.}
	\label{fig:simExpN50}
\end{figure*}

\paragraph{Additional performance metrics}
In \cref{fig:simGaussN500,fig:simExpN500,fig:simGaussN50,fig:simExpN50}, we plot four additional performance metrics to complement the F1 scores from \cref{fig:simulationsF1} in the main part of the paper. The metrics are standard and are defined as follows:
\begin{enumerate}[itemsep=-2pt,before=\vspace{-1em},after=\vspace{-1em},label={$\bullet$}]
\item True positive rate (TPR): number of correctly-identified edges divided by the number of edges in the ground-truth graph.
\item False discovery rate (FDR): number of incorrectly-identified edges divided by the number of edges in the estimated graph.
\item Structural Hamming distance (SHD): number of changes (i.e., edge removals, edge additions, and edge reversals) required to go from one (unweighted) graph to another.
\item Frobenius norm (FRO) of the difference between two weighted matrices (i.e., between a ground-truth adjacency matrix and an estimated matrix).
\end{enumerate}
Note that the Frobenius norm does not apply to the tsGFCI algorithm, which only returns unweighted edges.

For $n=500$, DYNOTEARS generally outperforms the other algorithms for both Gaussian (\cref{fig:simGaussN500}) and exponential noise (\cref{fig:simExpN500}); there are some exceptions to this when the number of variables is small, $d \in \{5,10,20\}$. For $n=50$ (see \cref{fig:simGaussN50,fig:simExpN50}), NOTEARS + Lasso and LiNGAM output estimated graphs that are significantly denser than the ground truth. As a result, while these two algorithms have large TPRs, their overall performance is not competitive due to a large number of false positives.

\FloatBarrier

\section{S\&P100 application}

\subsection{Parameter selection}
\begin{figure}[ht!]
\includegraphics[width=\columnwidth]{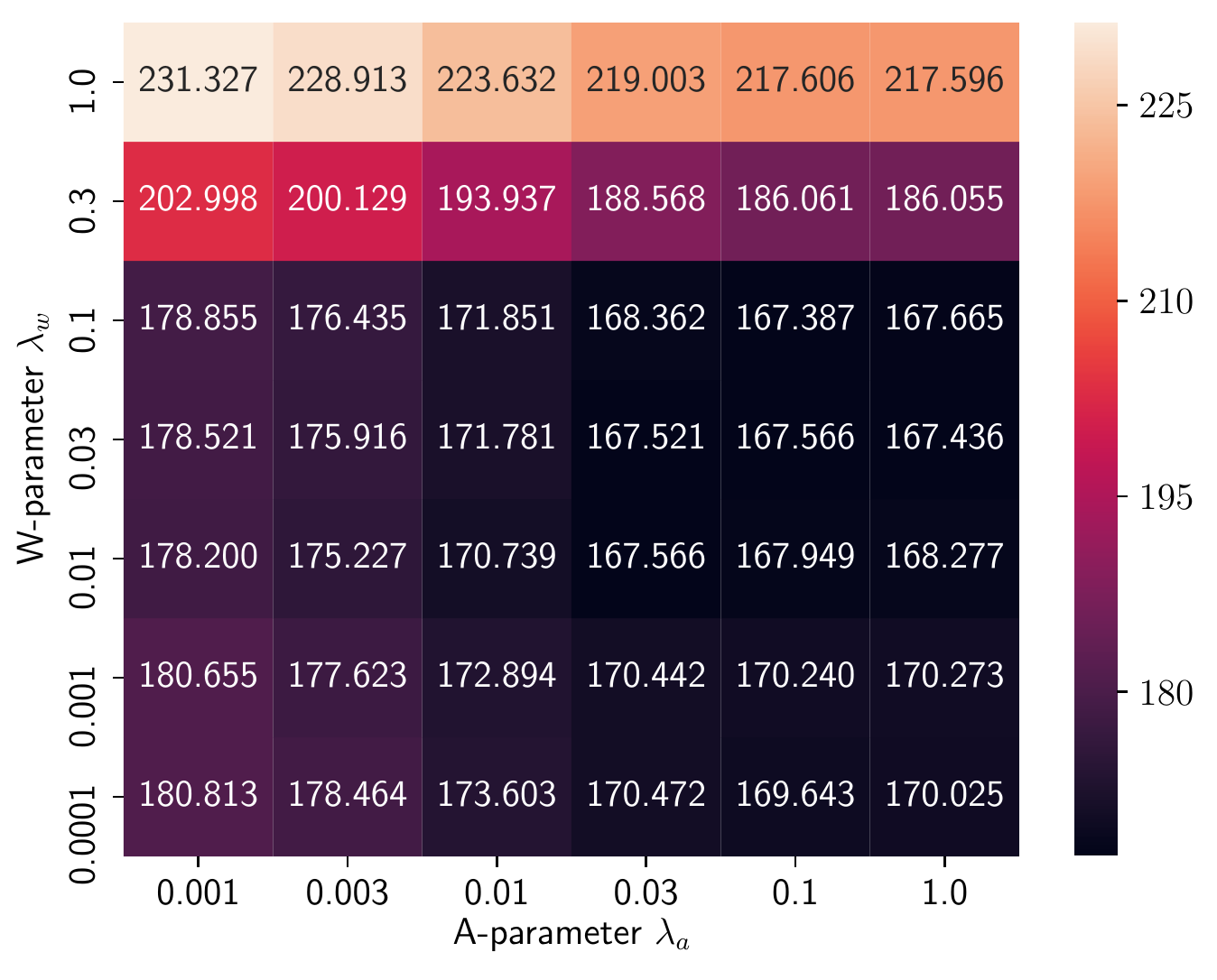}
\caption{Heatmap of validation loss using the Frobenius norm for the S\&P100 dataset across a range of $\lambda_{\Amat}$ and $\lambda_{\Wmat}$ parameters. Note that $\lambda_{\Wmat}=1$ and $\lambda_{\Amat}=1$ correspond to zero-matrices for $\Wmat$ and $\Amat$ and the loss is equal to the Frobenius norm of the dataset, $217.596$. The loss for static NOTEARS with $\lambda_{\Wmat}=0.1$ is $167.784$.}
\label{fig:validation_sp100}
\end{figure}

We hold out the last 400 trading days as a validation set and we discard 2 data points between the validation and trainining sets. This roughly corresponds to a 33\%/66\% split. We report the Frobenius norm across a range of parameter values in \cref{fig:validation_sp100}. Note that the multiples of $3$ are used as half-way points in the log-10 space and added to create a finer grid. However, we find that the surface is fairly smooth and we show these values for completeness. We select $\lambda_{\Wmat}=0.1$ and $\lambda_{\Amat}=0.1$, as they correspond to the smallest values in the grid.

\FloatBarrier

\section{DREAM4 application}\label{sec:DREAMsupplement}

\subsection{Cross validation}\label{sec:cvDream}

\begin{figure*}[ht!]
\includegraphics[width=\textwidth]{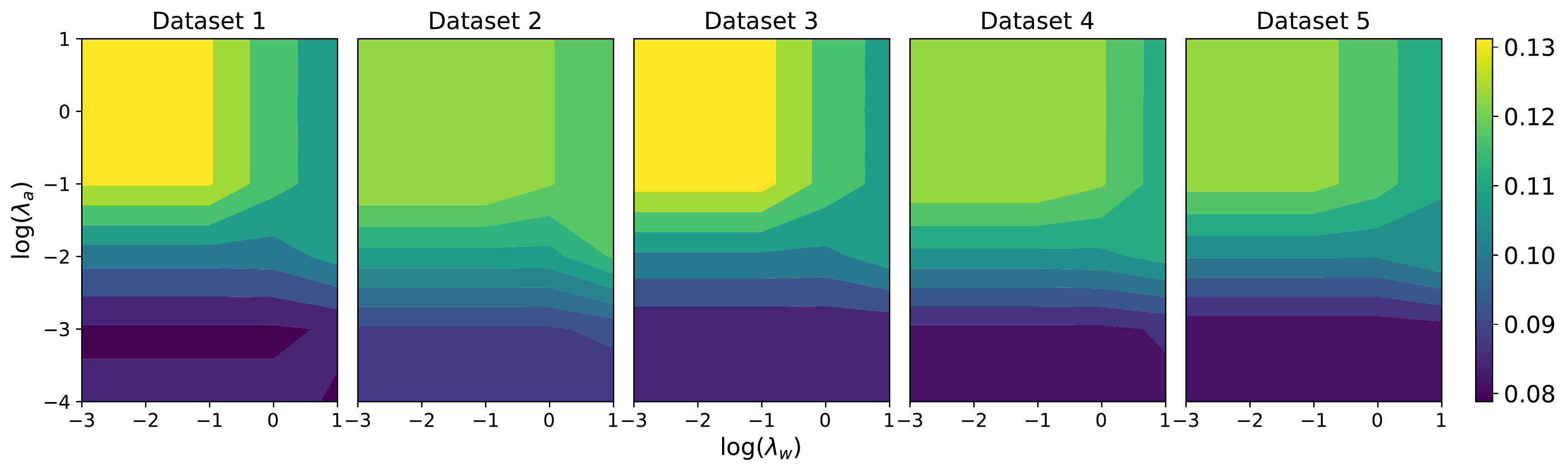}
\caption{Heatmaps of cross-validation RMSE for the 5 DREAM4 datasets across a range of $\lambda_{\Amat}$ and $\lambda_{\Wmat}$ parameters.}
\label{fig:cv}
\end{figure*}

We use $10$-fold cross validation to select the regularization parameters $\lambda_{\Wmat}$ and $\lambda_{\Amat}$. The number of folds is a natural choice, as each dataset consists of $10$ separate time series evaluated at $21$ time steps for the same $100$ genes. We evaluate performance on the validation set using the root mean squared error (RMSE). We find that for sufficiently low $\lambda_{\Amat}$, the RMSE is not sensitive to the values of $\lambda_{\Wmat}$
(see \Cref{fig:cv}). A plausible explanation is that our data consist of time steps separated at intervals of 50, but the underlying process is at a slower scale; the model therefore predominantly captures lagged inter-slice directed edges. As such, we set $\lambda_{\Wmat}$ to the largest value in the range of optimal RMSE to enforce sparsity and simplicity. We apply cross validation separately for each of the $5$ datasets in DREAM4, and we use the optimal parameters in each case to compute the average AUPR and AUROC. %

\subsection{Comparison to other methods}

In \cref{table:auroc,table:aupr}, we compare the performance of DYNOTEARS to that of other methods. We obtain performance metrics for other algorithms from \citet{lu2019causal}.

The performance of our method can be compared to other solvers. Note that our model returns two matrices, $\Wmat$ and $\Amat$, which can be interpreted as learning fast-acting ($\Wmat$) and slow-acting ($\Amat$) influences. As such, we combine the two matrices by an element-wise sum to generate our final weight matrix. The results are summarized in \Cref{table:auroc} and \Cref{table:aupr}. Note that there are missing values in the table because they were not initially reported.

\begin{table*}[ht!]
	\centering
	\resizebox{\textwidth}{!}{
	\begin{tabular}{p{2.5cm}|*{10}{|c}}

\toprule

  Algorithm & Method &  Average &  STD &  Network &  Network  &  Network &  Network &  Network &  Overall &  DBN \\
 {} & Type &  AUROC &  AUROC &  1 &  2 &  3 &  4 &  5 &  Rank &  Rank \\

\midrule
  \textbf{DYNOTEARS }&        \textbf{DBN} &         \textbf{0.664} &      \textbf{0.047} &      \textbf{0.748} &      \textbf{0.612} &     \textbf{0.634} &      \textbf{0.674} &      \textbf{0.653} &            \textbf{8} &       \textbf{2} \\
    Ebdbnet &         DBN &          0.643 &            &            &            &            &            &            &            11 &       4 \\
      G1DBN &         DBN &          0.676 &      0.030 &      0.680 &      0.640 &      0.680 &      0.660 &      0.720 &             6 &       1 \\
    ScanBMA &         DBN &          0.657 &            &            &            &            &            &            &            10 &       3 \\
     VBSSMa &         DBN &          0.624 &      0.060 &      0.590 &      0.560 &      0.590 &      0.670 &      0.710 &            13 &       5 \\
     VBSSMb &         DBN &          0.618 &      0.060 &      0.560 &      0.570 &      0.620 &      0.640 &      0.700 &            14 &       6 \\
      Jump3 &          DT &          0.720 &      0.040 &      0.770 &      0.670 &      0.740 &      0.680 &      0.740 &             2 &           \\
       CSIc &          GP &          0.610 &      0.030 &      0.650 &      0.560 &      0.630 &      0.610 &      0.600 &            15 &           \\
       CSId &          GP &          0.728 &      0.010 &      0.740 &      0.710 &      0.720 &      0.740 &      0.730 &             1 &           \\
     GP4GRN &          GP &          0.686 &      0.040 &      0.720 &      0.620 &      0.700 &      0.700 &      0.690 &             4 &           \\
     ARACNE &          MI &          0.558 &      0.010 &      0.560 &      0.540 &      0.560 &      0.550 &      0.580 &            18 &           \\
       CLR &          MI &          0.678 &      0.030 &      0.700 &      0.630 &      0.710 &      0.670 &      0.680 &             5 &           \\
      MRNET &          MI &          0.672 &      0.030 &      0.680 &      0.630 &      0.710 &      0.660 &      0.680 &             7 &           \\
       TSNI &         ODE &          0.566 &      0.030 &      0.550 &      0.550 &      0.600 &      0.540 &      0.590 &            17 &           \\
      BETS &         VAR &          0.688 &      0.060 &      0.780 &      0.650 &      0.640 &      0.700 &      0.670 &             3 &           \\
       Enet &         VAR &          0.662 &      0.050 &      0.730 &      0.620 &      0.620 &      0.670 &      0.670 &             9 &           \\
       GCCA &         VAR &          0.584 &      0.020 &      0.600 &      0.570 &      0.600 &      0.580 &      0.570 &            16 &           \\
      LASSO &         VAR &          0.643 &            &            &            &            &            &            &            11 &           \\
\bottomrule

\end{tabular}
}
  \caption{AUROC scores of 18 structure-learning algorithms on the DREAM4 gene-expression dataset. Values for methods other than DYNOTEARS are from \citet{lu2019causal}.}
  \label{table:auroc}
\end{table*}

\begin{table*}[ht!]
	\centering
	\resizebox{\textwidth}{!}{
	\begin{tabular}{p{2.5cm}|*{10}{|c}}

\toprule

  Algorithm & Method &  Average &  STD &  Network &  Network  &  Network &  Network &  Network &  Overall &  DBN \\
 {} & Type &  AUPR &  AUPR &  1 &  2 &  3 &  4 &  5 &  Rank &  Rank \\

\midrule
    \textbf{DYNOTEARS} &         \textbf{DBN} &           \textbf{0.173} &      \textbf{0.041} &      \textbf{0.235} &      \textbf{0.110} &      \textbf{0.177} &      \textbf{0.188} &      \textbf{0.155} &     \textbf{4} &       \textbf{1} \\
      Ebdbnet &         DBN &           0.043 &             &            &            &            &            &            &    21 &       6 \\
        G1DBN &         DBN &           0.110 &       0.010 &      0.110 &      0.100 &      0.130 &      0.100 &      0.110 &     8 &       2 \\
      ScanBMA &         DBN &           0.101 &             &            &            &            &            &            &     9 &       3 \\
       VBSSMa &         DBN &           0.086 &       0.020 &      0.080 &      0.050 &      0.110 &      0.100 &      0.090 &    12 &       5\\
       VBSSMb &         DBN &           0.096 &       0.030 &      0.090 &      0.060 &      0.120 &      0.120 &      0.090 &    11 &       4\\
    dynGENIE3 &          DT &           0.198 &       0.050 &      0.220 &      0.140 &      0.250 &      0.220 &      0.160 &     2 &           \\
       GENIE3 &          DT &           0.072 &       0.020 &      0.050 &      0.060 &      0.100 &      0.060 &      0.090 &    14 &           \\
        Jump3 &          DT &           0.182 &       0.050 &      0.260 &      0.110 &      0.190 &      0.170 &      0.180 &     3 &           \\
         CSIc &          GP &           0.070 &       0.040 &      0.130 &      0.030 &      0.070 &      0.070 &      0.050 &    16 &           \\
        CSId &          GP &           0.208 &       0.030 &      0.260 &      0.170 &      0.220 &      0.200 &      0.190 &     1 &           \\
       GP4GRN &          GP &           0.162 &       0.050 &      0.220 &      0.100 &      0.160 &      0.210 &      0.120 &     6 &           \\
       ARACNE &          MI &           0.046 &       0.010 &      0.030 &      0.040 &      0.060 &      0.040 &      0.060 &    20 &           \\
          CLR &          MI &           0.072 &       0.020 &      0.050 &      0.060 &      0.110 &      0.060 &      0.080 &    14 &           \\
        MRNET &          MI &           0.068 &       0.020 &      0.040 &      0.060 &      0.100 &      0.060 &      0.080 &    18 &           \\
       tl-CLR &          MI &           0.168 &       0.050 &      0.180 &      0.110 &      0.240 &      0.150 &      0.160 &     5 &           \\
  Inferelator &         ODE &           0.069 &       0.010 &      0.063 &      0.071 &      0.075 &      0.073 &      0.062 &    17 &           \\
         TSNI &         ODE &           0.026 &       0.010 &      0.020 &      0.030 &      0.030 &      0.020 &      0.030 &    23 &           \\
         BETS &         VAR &           0.128 &       0.020 &      0.160 &      0.100 &      0.130 &      0.140 &      0.110 &     7 &           \\
         Enet &         VAR &           0.098 &       0.020 &      0.120 &      0.080 &      0.100 &      0.110 &      0.080 &    10 &           \\
         GCCA &         VAR &           0.050 &       0.020 &      0.040 &      0.040 &      0.070 &      0.070 &      0.030 &    19 &           \\
        LASSO &         VAR &           0.073 &             &            &            &            &            &            &    13 &           \\
  OKVAR-Boost &         VAR &           0.034 &       0.020 &      0.050 &      0.050 &      0.030 &      0.020 &      0.020 &    22 &           \\
\bottomrule

\end{tabular}
}
  \caption{AUPR scores of 23 structure-learning algorithms on the DREAM4 gene-expression dataset. Values for methods other than DYNOTEARS are from \citet{lu2019causal}.}
  \label{table:aupr}
\end{table*}

\clearpage
\onecolumn %
\twocolumn %

\end{appendices}

\end{document}